# Medical Concept Representation Learning from Electronic Health Records and its Application on Heart Failure Prediction

Edward Choi*,MS; Andy Schuetz**,PhD; Walter F. Stewart**,PhD; Jimeng Sun*,PhD

*Georgia Institute of Technology, Atlanta, USA

**Research Development & Dissemination, Sutter Health, Walnut Creek, USA

Corresponding Author: Jimeng Sun

Georgia Institute of Technology

266 Ferst Drive, Atlanta, GA 30313

Tel: 404.894.0482

E-mail: jsun@cc.gatech.edu

Keywords:

Neural Networks, Representation learning, Predictive modeling, Heart Failure prediction.

## ABSTRACT

**Objective:** To transform heterogeneous clinical data from electronic health records into clinically meaningful constructed features using data driven method that rely, in part, on temporal relations among data.

**Materials and Methods:** The clinically meaningful representations of medical concepts and patients are the key for health analytic applications. Most of existing approaches directly construct features mapped to raw data (e.g., ICD or CPT codes), or utilize some ontology mapping such as SNOMED codes. However, none of the existing approaches leverage EHR data directly for learning such concept representation. We propose a new way to represent heterogeneous medical concepts (e.g., diagnoses, medications and procedures) based on co-occurrence patterns in longitudinal electronic health records. The intuition behind the method is to map medical concepts that are co-occuring closely in time to similar *concept vectors* so that their distance will be small. We also derive a simple method to construct *patient vectors* from the related medical concept vectors.

**Results:** For qualitative evaluation, we study similar medical concepts across diagnosis, medication and procedure. In quantitative evaluation, our proposed representation significantly improves the predictive modeling performance for onset of heart failure (HF), where classification methods (e.g. logistic regression, neural network, support vector machine and K-nearest neighbors) achieve up to 23% improvement in area under the ROC curve (AUC) using this proposed representation.

**Conclusion:** We proposed an effective method for patient and medical concept representation learning. The resulting representation can map relevant concepts together and also improves predictive modeling performance.

## Introduction

Growth in use of electronic health records (EHR) in health care delivery is opening unprecedented opportunities to predict patient risk, understand what works best for a given patient, and to personalize clinical decision-making. But, raw EHR data, represented by a heterogeneous mix of elements (e.g., clinical measures, diagnoses, medications, procedures) and voluminous unstructured content, may not be optimal for analytic uses or even for clinical care. While higher order clinical features (e.g., disease phenotypes) are intuitively more meaningful and can reduce data volume, they may fail to capture meaningful information inherent to patient data. We explored whether novel data driven methods that rely on the temporal occurrence of EHR data elements could yield higher order intuitively interpretable features that both capture pathophysiologic relations inherent to data and improve performance of predictive models.

Growth in use of EHRs is raising fundamental questions on optimal ways to represent structured and unstructured data. Medical ontologies such as SNOMED, RxNorm and LOINC offer structured hierarchical means of compressing data and of understand relations among data from different domains (e.g., disease diagnosis, labs, prescriptions). But, these ontologies do not offer the means of extracting meaningful relations inherent to longitudinal patient data. Scalable methods that can detect pathophysiologic relations inherent to longitudinal EHR data and construct intuitive features may accelerate more effective use of EHR data in clinical care and advances in performance of predictive analytics.

The abstract concepts inherent to existing ontologies does not provide a means to connect elements in different domains to a common underlying pathophysiologic constructs that are represented by how data elements co-occur in time. The data driven approach we developed logically organizes data into higher order constructs. Heterogeneous medical data were mapped to a low-dimensional space that accounted for temporal clustering of similar concepts (e.g., A1c lab test, ICD-9 code for diabetes, prescription for metformin). Co-occurring clusters (e.g., diabetes and peripheral neuropathy) were then identified and formed into higher order pathophysiologic feature sets organized by prevalence.

We propose to learn such a medical concept representation on longitudinal EHR data based on a state-of-the-art neural network model. We also propose an efficient way to derive patient representation based on the medical concept representation (or *medical concept vectors*). We calculated for a set of diseases their closest diseases, medications and procedures to demonstrate the clinical knowledge captured by the medical concept representations. We use those learned representation for heart failure prediction tasks, where significant performance improvement up to 23% in AUC can be obtained on many classification models (logistic regression: AUC 0.766 to 0.791, SVM: AUC 0.736 to 0.791, neural network: AUC 0.779 to 0.814, KNN: AUC 0.637 to 0.785)

## BACKGROUND

### Representation Learning in Natural Language Processing

Recently, neural network based representation learning has shown success in many fields such as computer vision [1] [2] [3], audio processing [4] [5] and natural language processing (NLP) [6] [7] [8]. We discuss representation learning in NLP, in particular, as our proposed method is based on Skip-gram [6] [9], a popular method for learning word representations.

Mikolov et al. [6] [9] proposed Skip-gram, a simple model based on neural network that can learn real-valued multi-dimensional vectors that capture relations between words by training on massive amount of text. The trained real-valued vectors will have similar values for syntactically and semantically close words such as *dog* and *cat* or *would* and *could*, but distinct values for words that are not. Pennington et al. [10] proposed GloVe, a word representation algorithm based on the global co-occurrence matrix. While GloVe and Skip-gram essentially achieve the same goal by taking different approaches, GloVe is computationally faster than Skip-gram as it precomputes co-occurrence information before the actual learning. Skip-gram, however, requires less number of hyper-parameters to tune than GloVe, and generally shows better performance [11].

As natural language text can be seen as a sequence of codes, medical records such as diagnoses, medications and procedures can also be seen as a sequence of codes over time. In this work, we propose a framework for mapping raw medical concepts (e.g., ICD9, CPT) into related concept vectors using Skip-gram and validating the utility of the resulting medical concept vectors.

**Representation Learning in the Clinical Field**

A few researchers applied representation learning in the clinical field recently. Minnaro-Gimenez et al. [12] learned the representations of medical terms by applying Skip-gram to various medical text. They collected the medical text from PubMed, Merck Manuals, Medscape and Wikipedia. De Vine et al. [13] learned the representations of UMLS concepts from free-text patient records and medical journal abstracts. The first pre-processed the text to map the words to UMLS concepts, then applied Skip-gram to learn the representations of the concepts.

More recently, Choi et al. [14] applied Skip-gram to structured dataset from a health insurance company, where the dataset consisted of patient visit records along with diagnosis codes(ICD9), lab test results(LOINC), and drug usage(NDC). Their goal, to learn efficient representations of medical concepts, partially overlaps with our goal. Our study however, is focused on learning the representations of medical concepts and using them to generate patient representations, apply them to a real-world prediction problem to demonstrate improved performance provided by the efficient representation learning.

## MATERIALS AND METHODS

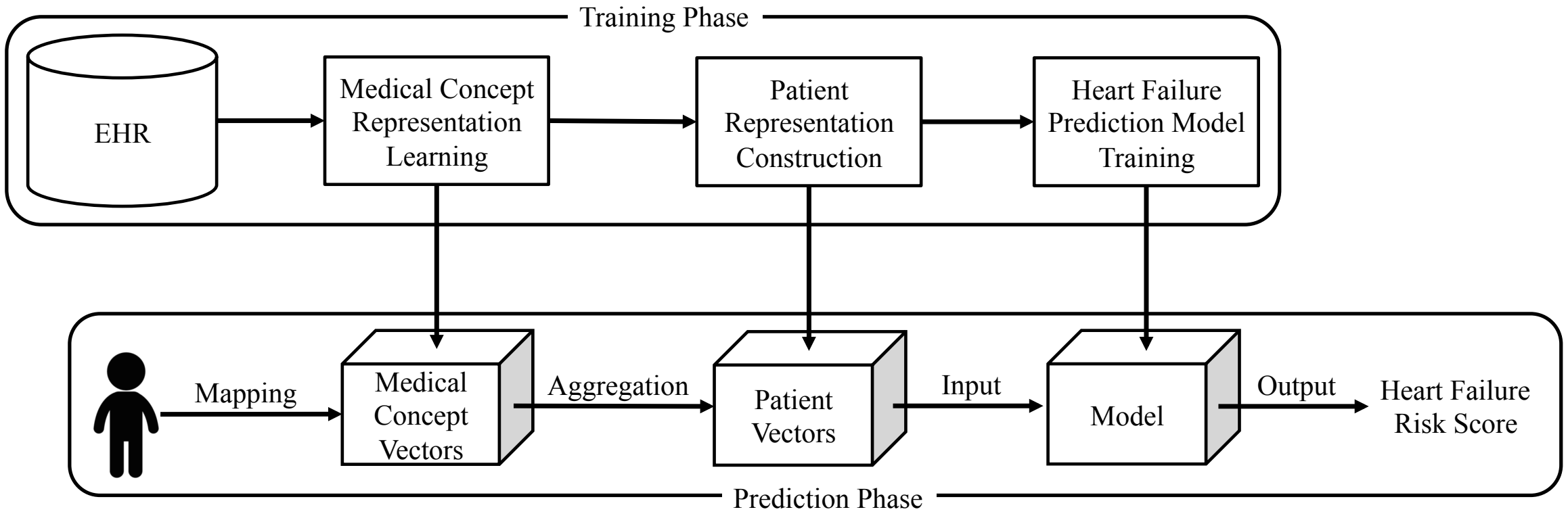

**Figure 1.** Flowchart of the proposed method

In Figure 1, we give a high-level overview of the steps we take to perform HF prediction. In the training phase, we first train medical concept vectors from the EHR dataset using Skip-gram. Then, we construct patient representation using the medical concept vectors. The patient representation is then used to train heart failure prediction models using various classifiers, namely logistic regression, support vector machine (SVM), multi-layer perceptron with one hidden layer (MLP) and K-nearest neighbors classifier (KNN). In the prediction phase, we map the medical record of a patient to medical concept vectors and generate patient vectors by aggregating the concept vectors. Then we plug the patient vectors into the trained model, which in turn will generate the risk score for heart failure.

In the following sections, we will describe medical concept representation learning and patient representation construction in more detail.

**Medical Concept Representation Learning**

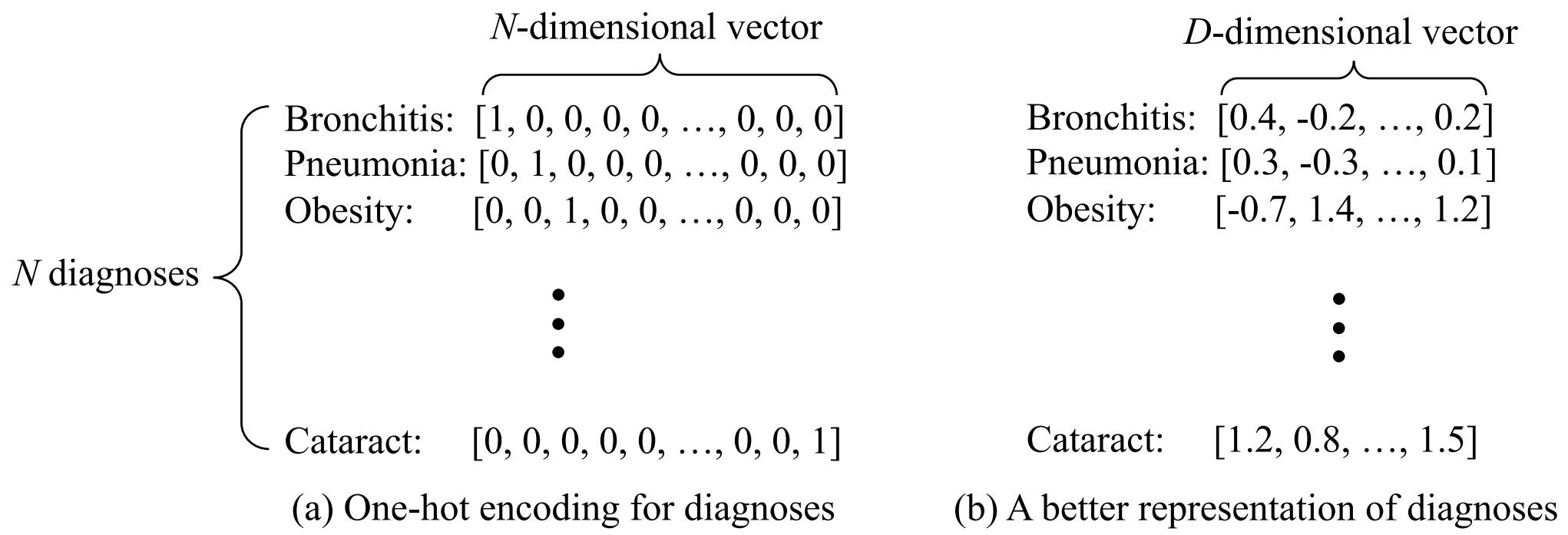

**Figure 2.** Two different representation of diagnoses. Typically, raw data dimensionality $N$(~10,000) is much larger than concept dimensionality $D$(50~1,000)

Figure 2 depicts a straightforward motivation for using a better representation for medical concepts. Figure 2(a) shows one-hot encoding of $N$ unique diagnoses using $N$-dimensional vectors. It is easy to see that this is not an effective representation in that the difference between *Bronchitis* and *Pneumonia* are the same as the difference between *Pneumonia* and *Obesity*. Figure 2(b) shows a better representation in that *Bronchitis* and *Pneumonia* share similar values compared to other diagnoses. By using Skip-gram, we will be able to better represent not only diagnoses but also medications and procedures as multi-dimensional real-valued vectors that will capture the latent relations between them.

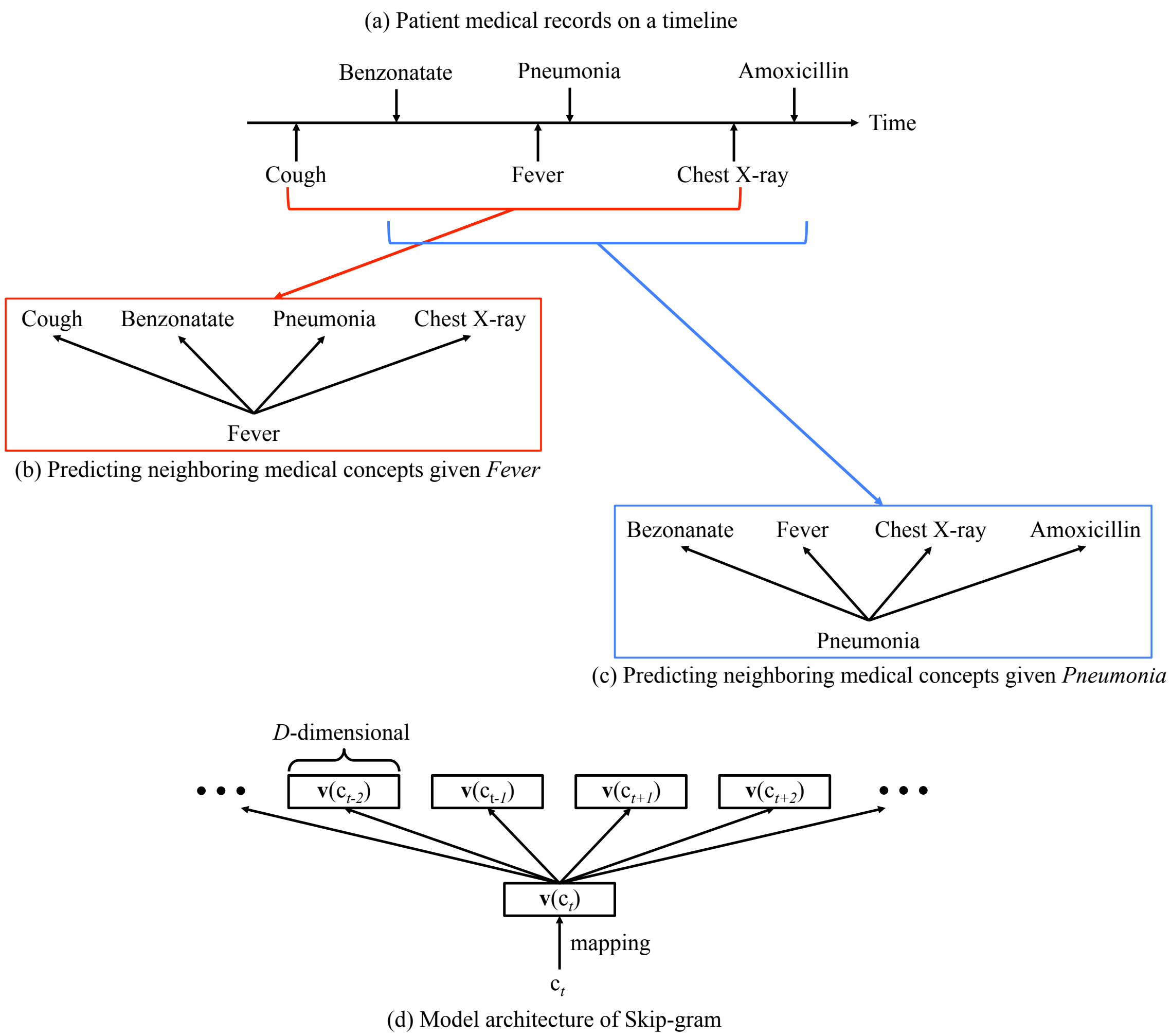

**Figure 3.** Training examples and the model architecture of Skip-gram

Figure 3(a) is an example of a patient medical record in a temporal order. Skip-gram assumes the meaning of a concept is determined by its context(or neighbors). Therefore, given a sequence of concepts, Skip-gram picks a target concept and tries to predict its neighbors, as shown by Figure 3(b). Then we slide the context window, pick the next target and do the same context prediction, which is shown by Figure 3(c). Since the goal of Skip-gram is to learn the vector representation of concepts, we need to convert medical concepts to *D*-dimensional vectors, where *D* is a user-chosen value typically between 50 and 1000.

Therefore the actual prediction is conducted with vectors, as shown by Figure 3(d), where $c_t$ is the concept at the $t$-th timestep, $\mathbf{v}(c_t)$ the vector that represents $c_t$. The goal of Skip-gram is to maximize the following average log probability,

$$\frac{1}{T}\sum_{t=1}^{T}\sum_{-w\le j\le w, j\ne 0}\log p\left(c_{t+j}|c_t\right), \text{where}$$

$$p\left(c_{t+j}|c_t\right) = \frac{exp\left(\mathbf{v}\left(c_{t+j}\right)^{\mathrm{T}}\mathbf{v}(c_t)\right)}{\sum_{c=1}^{N} exp\left(\mathbf{v}(c)^{\mathrm{T}}\mathbf{v}(c_t)\right)}$$

where $T$ is the length of the sequence of medical concepts, $w$ the size of the context window, $c_t$ the target medical concept at timestep $t$, $c_{t+j}$ the neighboring medical concept at timestep $t+j$, $\mathbf{v}(c)$ the vector that represents the medical concept $c$, $N$ the total number of medical concepts. The size of the context window is typically set to 5, giving us 10 concepts surrounding the target concept. Note that the conditional probability is expressed as a softmax function. Simply put, by maximizing the softmax score of the inner product of the neighboring concepts, Skip-gram learns real-valued vectors that efficiently capture the fine-grained relations between concepts. It needs to be mentioned that our formulation of Skip-gram is different from the original Skip-gram. In Mikolov et al. [9], they distinguish the vectors for the target concept and the vectors for the neighboring concept. In our formulation, we force the two sets of vectors to hold the same values as suggested by [15]. This simpler formulation allowed faster training and impressive results.

**Patient Representation Construction**

In this section, we describe a simple derivation of patient representation using the learned medical concept vectors. One of the impressive features of Skip-gram in Mikolov et al. [9] was that the word vectors supported syntactically and semantically meaningful linear operations that enabled word analogy calculations such that the resulting vector of King – Man + Woman is closest to Queen vector.

We expect that the medical concept representations learned by Skip-gram will show similar properties so that the concept vectors will support clinically meaningful vector additions. Then, an efficient representation of a patient will be as simple as converting all medical concepts in his medical history to medical concept vectors, then summing all those vectors to obtain a single representation vector, as shown in Figure 4. In the experiments, we show examples of clinically meaningful concept vector additions.

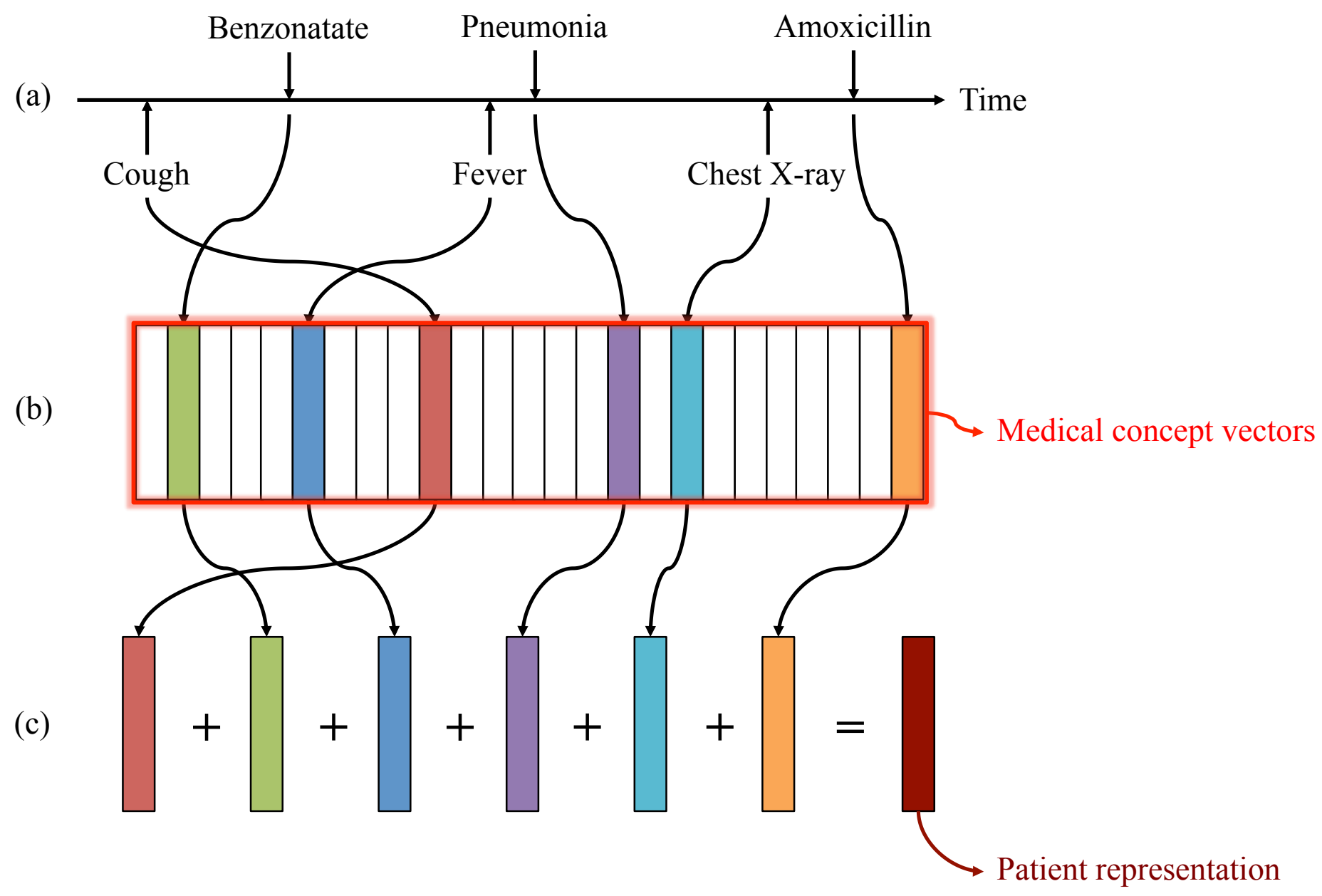

**Figure 4.** Patient representation construction. (a) represents a medical record of a patient on a timeline. (b) The medical concepts are represented as vectors using the trained medical concept vectors. (c) The patient is represented as a vector by summing all medical concept vectors.

## EXPERIMENTS AND RESULTS

### Population and Source of Data

Data were from Sutter Palo Alto Medical Foundation (Sutter-PAMF) primary care patients. Sutter-PAMF is a large primary care and multispecialty group practice that has used an EHR for more than a decade. The study dataset was extracted with cases and controls identified within the interval from 05/16/2000 to 05/23/2013. The EHR data included demographics, smoking and alcohol consumptions, clinical and laboratory values, International Classification of Disease version 9 (ICD-9) codes associated with encounters, order, and referrals, procedure information in Current Procedural Terminology (CPT) codes and medication prescription information in medical names. The dataset contained 265,336 patients with 555,609 unique clinical events in total.

**Configuration for Medical Concept Representation Learning**

To apply Skip-gram, we scanned through encounter, medication order, procedure order and problem list records of all 265,336 patients, and extracted diagnosis, medication and procedure codes assigned to each patient in temporal order. If a patient received multiple diagnoses, medications or procedures at a single visit, then those medical codes were given the same timestamp. The respective number of unique diagnoses, medications and procedures was 11,460, 17,769 and 9,370 totaling to 38,599 unique medical concepts. We used 100-dimensional vectors to represent medical concepts(i.e. $D$=100 in Figure 2(b)), considering 300 was sufficient to effectively represent 692,000 vocabularies in NLP. [9]

We used Theano [16], a Python library for evaluating mathematical expression to implement Skip-gram. Theano can also take advantage of GPUs to greatly improve the speed of calculations involving large matrices. For optimization, we used Adadelta [17], which employs adaptive learning rate. Unlike stochastic gradient descent (SGD), which is widely used for training neural networks, Adadelta does not depend very strongly on the setting of the learning rate, and shows good performance. Using Theano 0.7 and CUDA 7 on an Ubuntu machine with Xeon E5-2697 and Nvidia Tesla K80, it took approximately 43 hours to run 10 epochs of Adadelta with the batch size of 100.

## Evaluation of Medical Concept Representation Learning

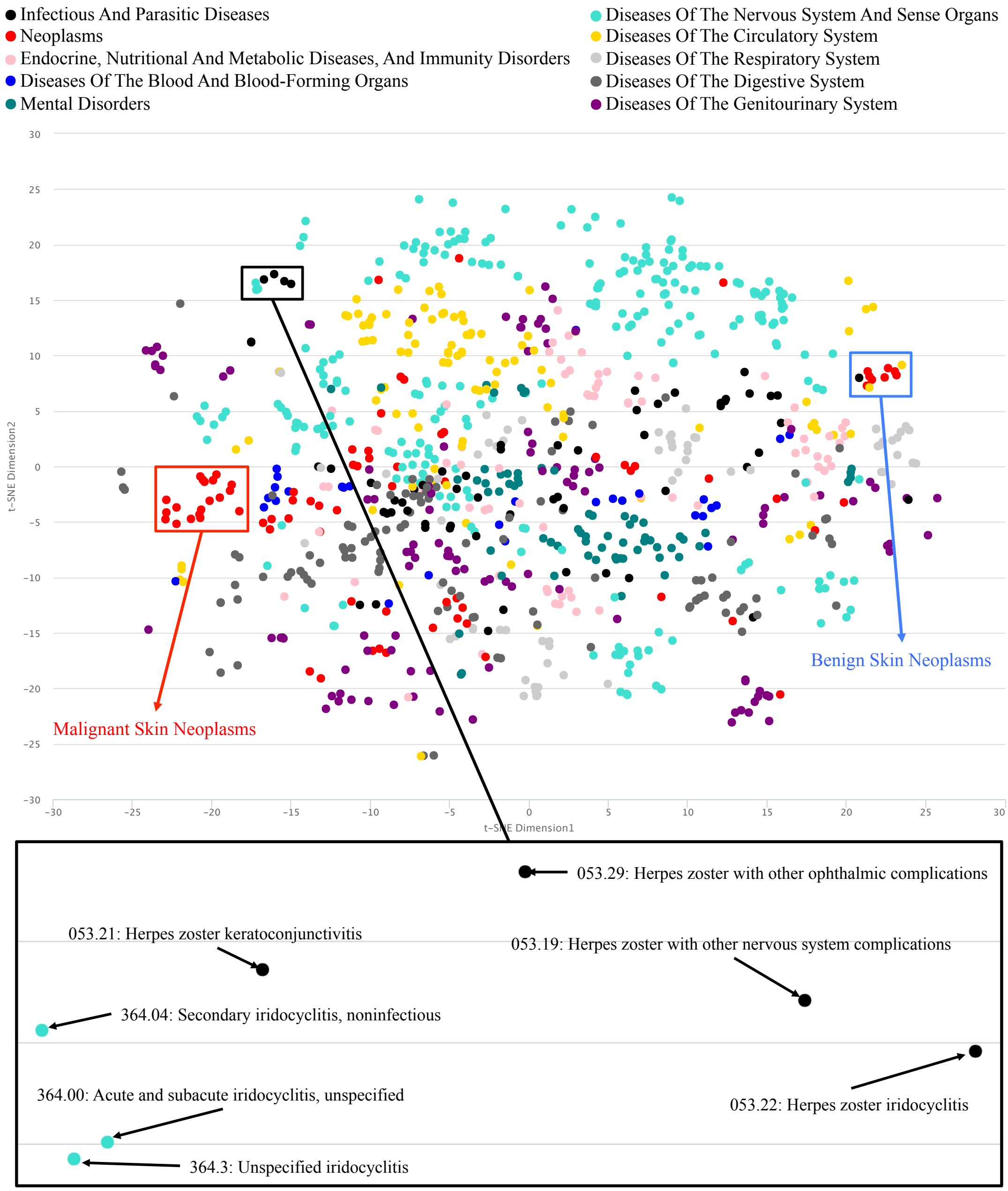

**Figure 5.** Diagnosis vectors projected to a 2D space by t-SNE

Figure 5 shows the trained diagnosis vectors plotted in a 2D space, where we used t-SNE [18] to reduce the dimensions from 100 to 2. t-SNE is a dimensionality reduction algorithm that was specifically developed for plotting high-dimensional data into a two or three dimensional space. We randomly chose 1,000 diagnoses from 10 uppermost categories of ICD-9, which are displayed at the top of the figure. It is readily visible that diagnoses are generally well grouped by their corresponding categories. However, if diagnoses from the same category are in fact quite different, they should be apart. This is shown by the red box and the blue box in Figure 5. Even though they are from the same *neoplasms* category, red box indicates a group of *malignant skin neoplasms* (172.X, 173.X) while blue box indicates a group of *benign skin neoplasms* (216.X). Detailed figure of the red and blue boxes are in the supplementary section. What is more, as the black box shows, diagnoses from different groups are located close to one another if they are actually related. In the black box, *iridocyclitis* and *eye infections related to herpes zoster* are closely located, which corresponds to the fact that approximately 43% *herpes zoster ophthalmicus* (HZO) patients develop *iridocyclitis*. [19]

In order to see how well the representation learning captured the relations between medications and procedures as well as diagnoses, we conduct the following study. We chose 100 diagnoses that occurred most frequently in the data, obtained for each diagnosis 50 closest vectors in terms of cosine similarity, picked 5 diagnosis, medication and procedure vectors among the 50 vectors. Table 2 depicts a portion of the entire lis. Note that some cells contain less than 5 items, which is because there was less than 5 items in the 50 closest vectors. The entire list is provided in the supplementary section.

**Table 1.** Examples of diagnoses and their closest medical concepts.

| | Diagnoses | Medications | Procedures |
|---|---|---|---|
| Acute upper respiratory infections (465.9) | -Bronchitis, not specified as acute or chronic (490)<br>-Cough (786.2)<br>-Acute sinusitis, unspecified (461.9)<br>-Acute bronchitis (466.0)<br>-Acute pharyngitis (462) | -Azithromycin 250 mg po tabs<br>-Promethazine-Codeine 6.25-10 mg/5ml po syrp<br>-Amoxicillin 500 mg po caps<br>-Fluticasone Propionate 50 mcg/act na susp<br>-Flonase 50 mcg/act na susp | -Pulse oximetry single<br>-Serv prov during reg sched eve/wkend/hol hrs<br>-Chest PA & lateral<br>-Gyn cytology (pap) pa<br>-Influenza vac (flu clinic only) 3+yo pa |
| Diabetes mellitus (250.02) | -Diabetes mellitus (250.00)<br>-Mixed hyperlipidemia (272.2)<br>-Other abnormal glucose (790.29)<br>-Obesity, unspecified (278.00)<br>-Pure hypercholesterolemia (272.0) | -Metformin hcl 500 mg po tabs<br>-Metformin hcl 1000 mg po tabs<br>-Glucose blood vi strp<br>-Lisinopril 10 mg po tabs<br>-Lisinopril 20 mg po tabs | -Diabetic eye exam (no bill)<br>-Diabetes education, int<br>-Ophthalmology, int<br>-Diabetic foot exam (no bill)<br>-Influenza vac 3+yr (v04.81) im |
| Edema (782.3) | -Anemia, unspecified (285.9)<br>-Congestive heart failure, unspecified (428.0)<br>-Unspecified essential hypertension (401.9)<br>-Atrial fibrillation (427.31)<br>-Chronic kidney disease, Stage III (moderate) (585.3) | -Furosemide 20 mg po tabs<br>-Hydrochlorothiazide 25 mg po tabs<br>-Hydrocodone-Acetaminophen 5-500 mg po tabs<br>-Cephalexin 500 mg po caps<br>-Furosemide 40 mg po tabs | -Debridement of nails, 6 or more<br>-OV est pt min serv<br>-EKG<br>-ECG and interpretation<br>-Chest PA & lateral |
| Tear film insufficiency, unspecified (375.15) | -Blepharitis, unspecified (373.00)<br>-Senile cataract, unspecified (366.10)<br>-Presbyopia (367.4)<br>-Preglaucoma, unspecified (365.00)<br>-Other chronic allergic conjunctivitis (372.14) | -Glasses<br>-Erythromycin 5 mg/gm op oint<br>-Patanol 0.1 % op soln | -Refraction<br>-Visual field exam extended<br>-Visual field exam limited<br>-Referral to ophthalmology, int<br>-Ophthalmology, int |
| Benign essential hypertension (401.1) | -Hyperlipidemia (272.4)<br>-Essential hypertension (401.9)<br>-Pure hypercholesterolemia (272.0)<br>-Mixed hyperlipidemia (272.2)<br>-Diabetes mellitus (250.00) | -Hydrochlorothiazide 25 mg po tabs<br>-Atenonol 50 mg po tabs<br>-Lisinopril 10 mg po tabs<br>-Lisinopril 40 mg po tabs<br>-Lisinopril 20 mg po tabs | -ECG and interpretation<br>-Influenza vac 3+yr im<br>-Immun admin im/sq/id/perc 1st vac only<br>-GI, int<br>-OV est pt lev 3 |

## Evaluation of Medical Concept Vector Additions

**Table 2.** Vector operations of medical vectors trained by Skip-gram.

| | Diagnoses | Medications | Procedures |
|---|---|---|---|
| Hypertension (401.9)<br>+<br>Obesity (278.0) | -Hyperlipidemia (272.4)<br>-Diabetes (250.00)<br>-Coronary atherosclerosis (414.00)<br>-Hypertension (401.1)<br>-Chronic kidney disease (585.3) | -Hydrochlorothiazide<br>-Valsartan<br>-Nifedipine<br>-Lisinopril<br>-Losartan potassium | N/A |
| Fever (780.60)<br>+<br>Cough (786.2) | -Pneumonia (486)<br>-Acute bronchitis (466.0)<br>-Acute upper respiratory infections (465.9)<br>-Bronchitis (490)<br>-Acute sinusitis (461.9) | -Azithromycin<br>-Promethazine-codeine<br>-Guaifenesin-codeine<br>-Proair HFA<br>-Levofloxacin | -X-ray chest<br>-Chest PA & Lateral<br>-Pulse oximetry<br>-Serv prov during reg sched eve/wkend/hol hrs<br>-Inhalation Rx for obstruction MDI/NEB |
| Visual Disturbance (368.8)<br>+<br>Pain in/around Eye (379.91) | -Tear film insufficiency (375.15)<br>-Visual discomfort (368.13)<br>-Regular astigmatism (367.21)<br>-Presbyopia (367.4)<br>-Blepharitis (373.00) | -Glasses<br>-Erythromycin ointment<br>-Patanol | -Ophthalmology<br>-Peripheral refraction<br>-Referral to ophthalmology<br>-Visual field exam<br>-Diabetic eye exam |
| Loss of Weight (783.21)<br>+<br>Anxiety State (300.00) | -Depressive disorder (311)<br>-Malais & fatigue (780.79)<br>-Insomnia (780.52)<br>-Generalized anxiety disorder (300.02)<br>-Esophageal reflux (530.81) | -Lorazepam<br>-Zolpidem tartrate<br>-Omeprazole<br>-Alprazolam<br>-Trazodone HCL | -Referral to GI<br>-ECG & Interpretation<br>-GI<br>-EKG<br>-Chest PA & Lateral |
| Hallucination (780.1)<br>+<br>Speech Disturbance (784.59) | -Dysarthria (784.51)<br>-Secondary parkinsonism (332.1)<br>-Senile dementia with delirium (290.3)<br>-Mental disorder (294.9)<br>-Paranoid state (297.9) | -Midorine HCL<br>-Risperdal<br>-Rivastigmine Tartrate<br>-Rivastigmine | -Referral to geriatrics<br>-Mental status exam<br>-Referral to neuropsychology<br>-Referral to speech therapy<br>Home visit est pt lev 2 |

Due to the difficulty of generating medically interesting examples, we chose 5 intuitive examples as shown by the first column of Table 3 to give a simple demonstration of the medical concept vector additions. We again generated 50 closest vectors to the sum of two medical concept vectors and picked 5 from each diagnosis, medication and procedure category.

## Setup for Heart Failure Prediction Evaluation

In this section, we first describe why we chose heart failure (HF) prediction task as an application. Then we briefly mention the models to use, followed by the description of the data processing steps to create the training data for all models. Lastly, the evaluation strategy will be followed by implementation details.

**Heart failure prediction task:** Onset of HF is associated with a high level of disability, health care costs and mortality (roughly ~50% risk of mortality within 5 years of diagnosis). [20] [21] There has been relatively little progress in slowing the progression of HF severity, largely because it is difficult to detect before actual diagnosis. As a consequence, intervention has primarily been confined to the time period after diagnosis, with little or no impact on disease progression. Earlier detection of HF could lead to improved outcomes through patient engagement and more assertive treatment with angiotensin converting enzyme (ACE)-inhibitors or Angiotensin II receptor blockers (ARBs), mild exercise, and reduced salt intake, and possibly other options [22] [23] [24] [25].

**Models for performance comparison:** We aim to emphasize the effectiveness of the medical concept representation and the patient representation derived from it. Therefore we trained four popular classifiers, namely logistic regression, MLP, SVM, and KNN using both one-hot vectors and medical concept vectors.

**Definition of Cases and Controls:** Criteria for incident onset of HF, are described in [26] and were adopted from [27]. The criteria are defined as: 1) Qualifying ICD-9 codes for HF appeared as a diagnosis code in either the encounter, the problem list, or the medication order fields. Qualifying ICD-9 codes are listed in the supplementary section. Qualifying ICD-9 codes with image and other related orders were excluded because these orders often represent a suspicion of HF, where the results are often negative; 2) a minimum of three clinical encounters with qualifying ICD-9 codes had to occur within 12 months of each other, where the date of diagnosis was assigned to the earliest of the three dates. If the time span between the first and second appearances of the HF diagnostic code was greater than 12 months, the date of the second encounter was used as the first qualifying encounter; 3) ages 50 or greater to less than 85 at the time of HF diagnosis.

Up to ten (nine on average) eligible primary care clinic-, sex-, and age-matched (in 5-year age intervals) controls were selected for each incident HF case. Primary care patients were eligible as controls if they had no HF diagnosis in the 12-month period before diagnosis of the incident HF case. Control subjects were required to have their first office encounter within one year of the matching HF case patient's first office visit, and have at least one office encounter 30 days before or any time after the case's HF diagnosis date to ensure similar duration of observations among cases and controls.

From 265,336 Sutter-PAMF patients, 3,884 incident HF cases and 28,903 control patients were identified.

**Data processing:** To train the four models, we generated the dataset again from the encounter, medication order, procedure order and problem list records of 3,884 cases and 28,903 controls. Based on the HF diagnosis date (HFDx) of each patient, we extracted all records from the 18-month period before the HFDx. To train the models with medical concept vectors, we converted the medical records to patient vectors as shown in Figure 4. To train the models with one-hot encoding, we converted the medical records to aggregated one-hot vectors in the same fashion as Figure 4, using one-hot vectors instead of medical concept vectors.

In order to study the relation between the medical concept vectors trained with different sizes of data and their influence on the models' prediction performance, we used three kinds of medical concept vectors: 1) The one trained with only HF cases (3,884 patients), 2) The one trained with HF cases and controls (32,787 patients), 3) The one trained with the full sample (265,336 patients). Note that medical concept vectors trained with smaller number of patients cover less number of medical concepts. Therefore, when converting patient records to patient vectors as Figure 4, we excluded all medical codes that did not have matching medical concept vectors. All input vectors were normalized to zero mean and unit variance.

**Evaluation strategy:** We used six-fold cross validation to train and evaluate all models, and to estimate how well the models will generalize to independent datasets. Prediction performance was measured using area under the ROC curve (AUC), on data not used in the training. We used the confidence score to calculate its AUC for SVM. Detailed explanation of the cross validation is given in the supplementary section.

**Implementation details:** Logistic regression and MLP were implemented with Theano and trained with Adadelta. SVM and KNN were implemented with Python Scikit-Learn. All models were trained by the same machine used for medical concept representation learning. Hyper-parameters used for training each model are described in the supplementary section.

**Evaluation of Heart Failure Prediction**

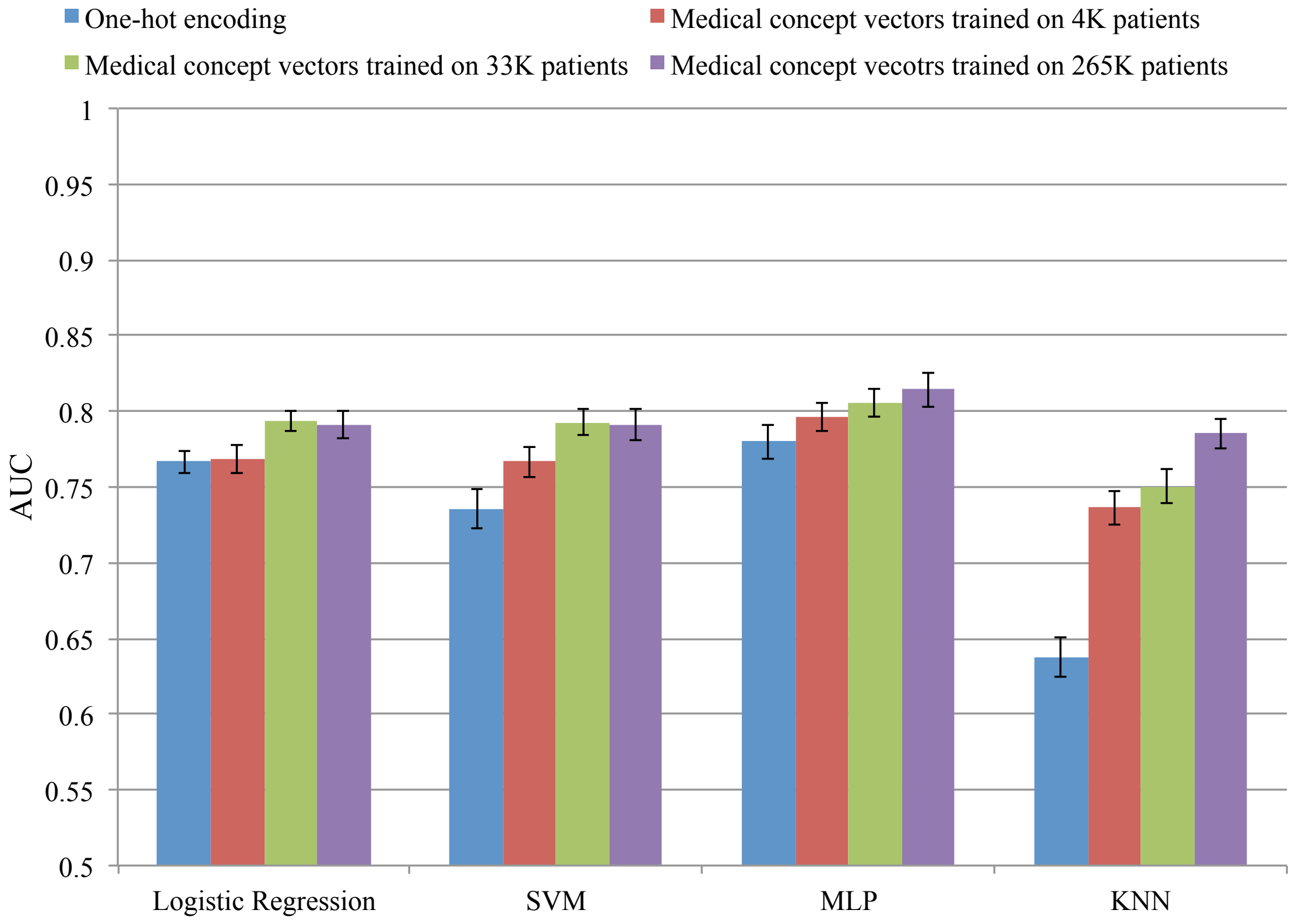

**Figure 6. H**eart failure prediction performance of various models and input vectors.

Figure 6 shows the average AUC of 6-fold cross validations of various models and input vectors. The colors represent different training input vectors. The error bars indicate the standard deviation derived from the 6-fold cross evaluation. The power of medical concept representation learning is evident as all models show significant improvement in the HF prediction performance. Logistic regression and SVM, both being linear models, show similar performance when trained with medical concept vectors, although SVM benefits slightly more from using the better representation of medical concepts. MLP also benefits from using medical concept vectors, and, being a non-linear model, shows better performance compared to logistic regression and SVM. It is interesting that KNN benefits the most from using the medical concept vectors, even the ones trained on the smallest dataset. Considering the fact that KNN classification is based on the distances between data points, this is a clear indication that proper medical concept representation can alleviate the sparsity problem induced by the simple one-hot encoding.

Figure 6 also tells us that medical concept representation is best learned with a large dataset as shown by Mikolov et al. [6] However, in most models, especially KNN, even the medical concept vectors trained with the smallest number of patients improves the prediction performance. It is quite surprising given the fact that we used less amount of information by excluding unmatched medical codes when using medical concept vectors trained with a small number of patients, the models still show better prediction performance. This again is a clear proof that medical concept representation learning provides more effective way to represent medical concepts than one-hot encoding.

**Table 3.** Training speed improvement. (*Since KNN does not require training, we display classification time instead)

| | Logistic regression | SVM | MLP | KNN* |
|---|---|---|---|---|
| One-hot encoding | 81.3 | 20.5 | 85.7 | 2900.82 |
| Medical concept vectors | 5.3 | 1.9 | 5.6 | 36.66 |
| Speed -up | x15.3 | x10.8 | x15.3 | x79.1 |

Table 4 depicts the training time for each model when using one-hot encoding and medical concept vectors. Considering the high-dimensionality of one-hot encoding, training the models with medical concept vectors should provide significant speed-up, as shown by the last row of Table 4. This shows that medical concept vectors not only improve performance, but also significantly reduce the training time.

Before discussing future work, we would like to emphasize the fact that our entire experiments were conducted completely without expert knowledge such as medical ontologies or features designed by medical experts. Using only the medical order records, we were able to produce clinically meaningful representation of medical concepts. This is an inspiring discovery that can be extended for numerous other medical problems.

**Future Work**

Although medical concept vectors have shown impressive results, it would be even more effective if deeper medical information could be embedded such as lab results or patient demographic information. This would enable us to represent the medical state of patients more accurately.

Using expert knowledge is another thing we should try. Even though we have shown impressive performance only by using medical records, this does not mean we cannot benefit from well-established expert medical knowledge, such as specific features or medical ontologies.

Another natural extension of our work is to address other medical problems. Although this work focused on the early detection of heart failure, our approach is very general that it could be applied to any kind of disease prediction problem. And the medical concept vectors can also be used in numerous medical applications as well.

**CONCLUSION**

We proposed a new way of representing heterogeneous medical concepts as real-valued vectors and constructing efficient patient representation using the state-of-the-art Deep Learning method. We have qualitatively shown that the trained medical concept vectors indeed captured medical insights compatible with our medical knowledge and experience. For the heart failure prediction task, medical concept vectors improved the performance of many classifiers, thus quantitatively proving its effectiveness. We discussed the limitation of our method and possible future works, which include deeper utilization of medical information, combining expert knowledge into our framework, and expanding our approach to various medical applications.

## SUPPLEMENTARY

**Table 4.** Qualifying ICD-9 codes for heart failure

| ICD-9 Code | Description |
|---|---|
| 398.91 | Rheumatic heart failure (congestive) |
| 402.01 | Malignant hypertensive heart disease with heart failure |
| 402.11 | Benign hypertensive heart disease with heart failure |
| 402.91 | Unspecified hypertensive heart disease with heart failure |
| 404.01 | Hypertensive heart and chronic kidney disease, malignant, with heart failure and with chronic kidney disease stage I through stage IV, or unspecified |
| 404.03 | Hypertensive heart and chronic kidney disease, malignant, with heart failure and with chronic kidney disease stage V or end stage renal disease |
| 404.11 | Hypertensive heart and chronic kidney disease, benign, with heart failure and with chronic kidney disease stage I through stage IV, or unspecified |
| 404.13 | Hypertensive heart and chronic kidney disease, benign, with heart failure and chronic kidney disease stage V or end stage renal disease |
| 404.91 | Hypertensive heart and chronic kidney disease, unspecified, with heart failure and with chronic kidney disease stage I through stage IV, or unspecified |
| 404.93 | Hypertensive heart and chronic kidney disease, unspecified, with heart failure and chronic kidney disease stage V or end stage renal disease |
| 428.0 | Congestive heart failure, unspecified |
| 428.1 | Left heart failure |
| 428.20 | Systolic heart failure, unspecified |
| 428.21 | Acute systolic heart failure |
| 428.22 | Chronic systolic heart failure |
| 428.23 | Acute on chronic systolic heart failure |
| 428.30 | Diastolic heart failure, unspecified |
| 428.31 | Acute diastolic heart failure |
| 428.32 | Chronic diastolic heart failure |
| 428.33 | Acute on chronic diastolic heart failure |
| 428.40 | Combined systolic and diastolic heart failure, unspecified |
| 428.41 | Acute combined systolic and diastolic heart failure |
| 428.42 | Chronic combined systolic and diastolic heart failure |
| 428.43 | Acute on chronic combined systolic and diastolic heart failure |
| 428.9 | Heart failure, unspecified |

**Red and Blue Box of Figure 5**

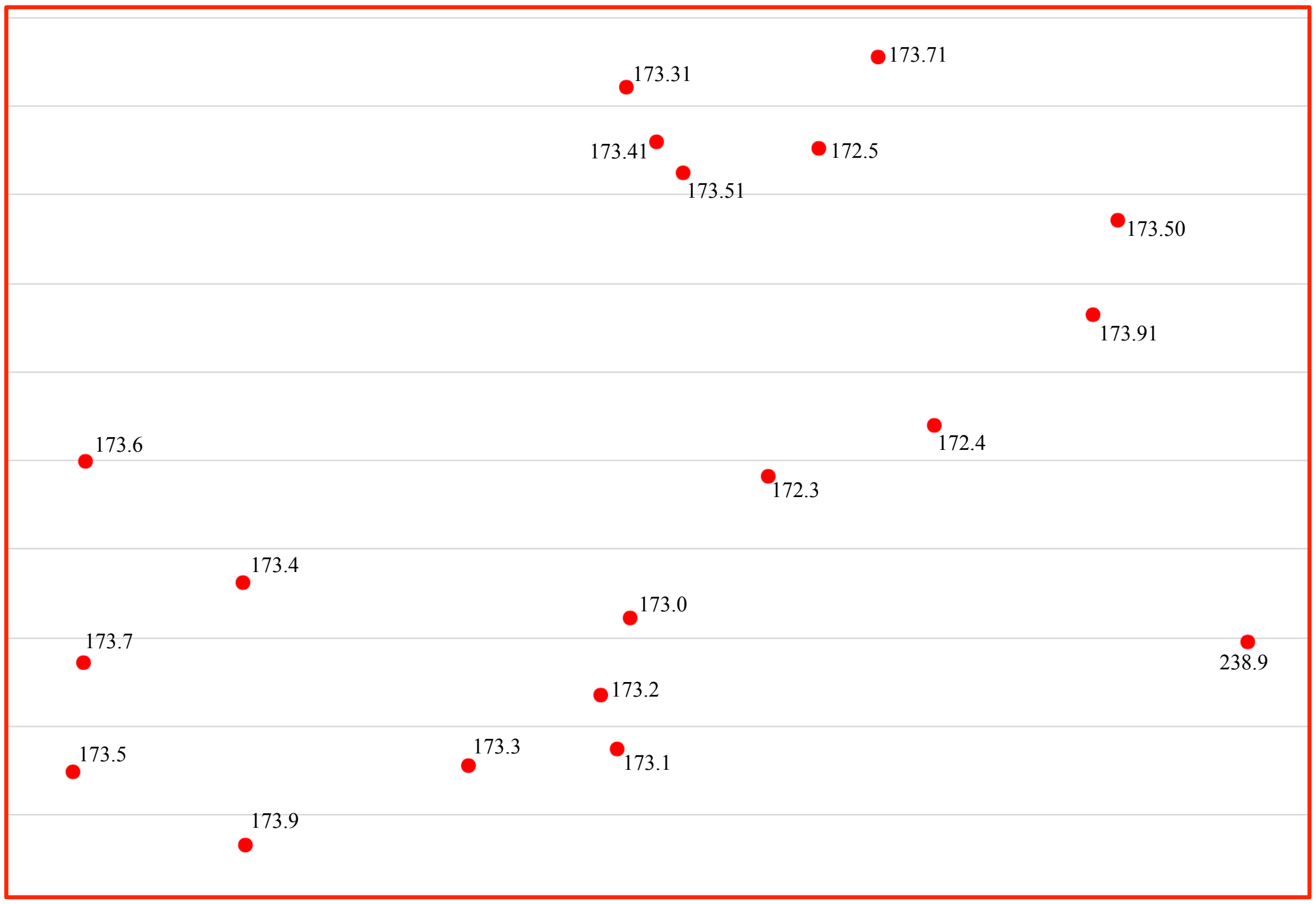

**Figure 7.** Detailed version of the red box of Figure 5

**Table 5.** List of ICD-9 codes that appear in Figure 7, and their descriptions

| ICD-9 Code | Description |
|---|---|
| 172.3 | Malignant melanoma of skin of other and unspecified parts of face |
| 172.4 | Malignant melanoma of skin of scalp and neck |
| 172.5 | Malignant melanoma of skin of trunk, except scrotum |
| 173.0 | Other and unspecified malignant neoplasm of skin of lip |
| 173.1 | Other and unspecified malignant neoplasm of skin of eyelid, including canthus |
| 173.2 | Other and unspecified malignant neoplasm of skin of ear and external auditory canal |
| 173.3 | Other and unspecified malignant neoplasm of skin of other and unspecified parts of face |
| 173.31 | Basal cell carcinoma of skin of other and unspecified parts of face |
| 173.4 | Other and unspecified malignant neoplasm of scalp and skin of neck |
| 173.41 | Basal cell carcinoma of scalp and skin of neck |
| 173.5 | Other and unspecified malignant neoplasm of skin of trunk, except scrotum |
| 173.50 | Unspecified malignant neoplasm of skin of trunk, except scrotum |
| 173.51 | Basal cell carcinoma of skin of trunk, except scrotum |
| 173.6 | Other and unspecified malignant neoplasm of skin of upper limb, including shoulder |
| 173.7 | Other and unspecified malignant neoplasm of skin of lower limb, including hip |
| 173.71 | Basal cell carcinoma of skin of lower limb, including hip |
| 173.9 | Other and unspecified malignant neoplasm of skin, site unspecified |
| 173.91 | Basal cell carcinoma of skin, site unspecified |
| 238.9 | Neoplasm of uncertain behavior, site unspecified |

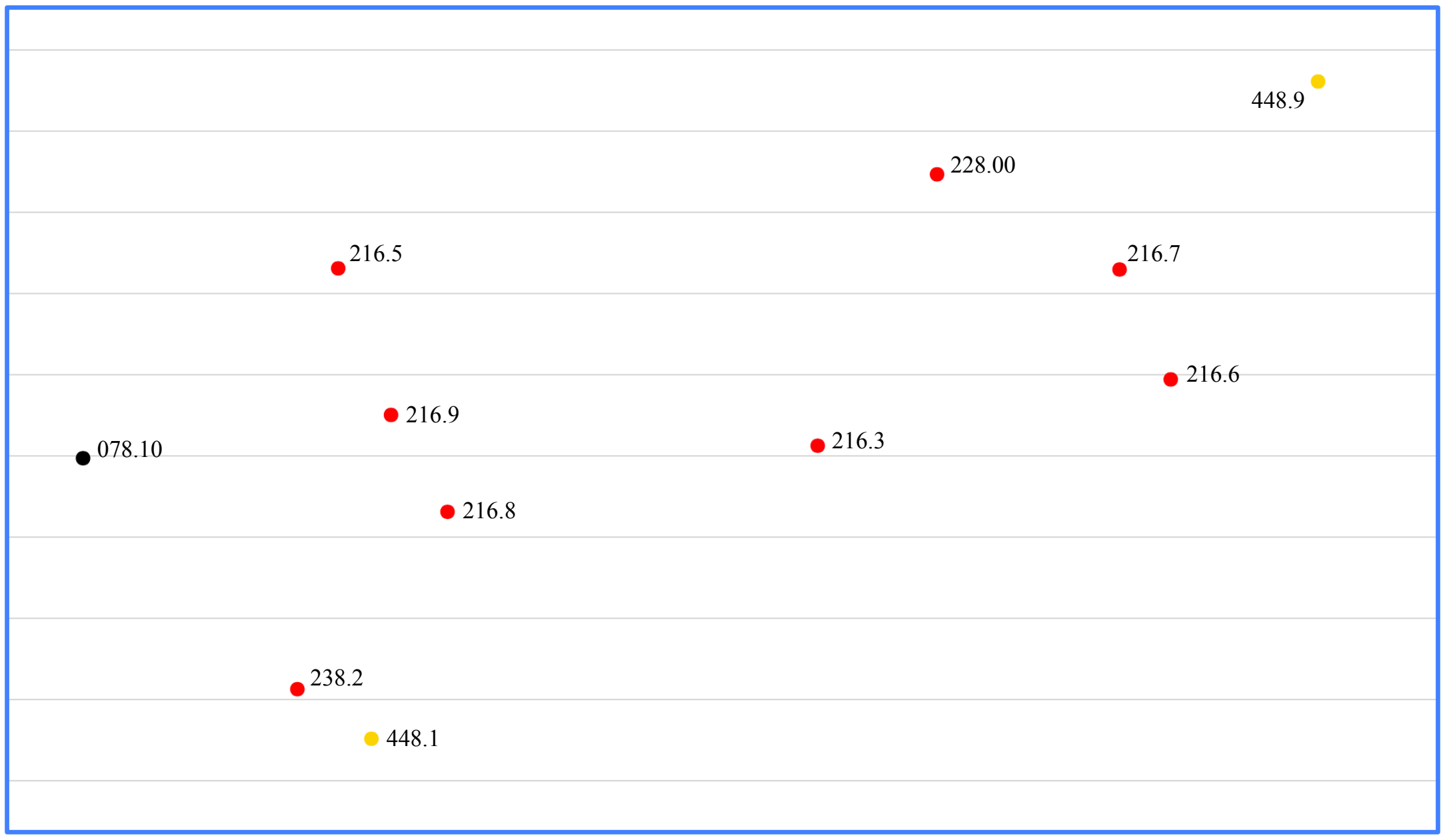

**Figure 8.** Detailed version of the blue box of Figure 5

**Table 6.** List of ICD-9 codes that appear in Figure 8, and their descriptions

| ICD-9 Code | Description |
|---|---|
| 078.10 | Viral warts, unspecified |
| 216.3 | Benign neoplasm of skin of other and unspecified parts of face |
| 216.5 | Benign neoplasm of skin of trunk, except scrotum |
| 216.6 | Benign neoplasm of skin of upper limb, including shoulder |
| 216.7 | Benign neoplasm of skin of lower limb, including hip |
| 216.8 | Benign neoplasm of other specified sites of skin |
| 216.9 | Benign neoplasm of skin, site unspecified |
| 228.00 | Hemangioma of unspecified site |
| 238.2 | Neoplasm of uncertain behavior of skin |
| 448.1 | Nevus, non-neoplastic |
| 448.9 | Other and unspecified capillary diseases |

**6-fold Cross Validation Scheme**

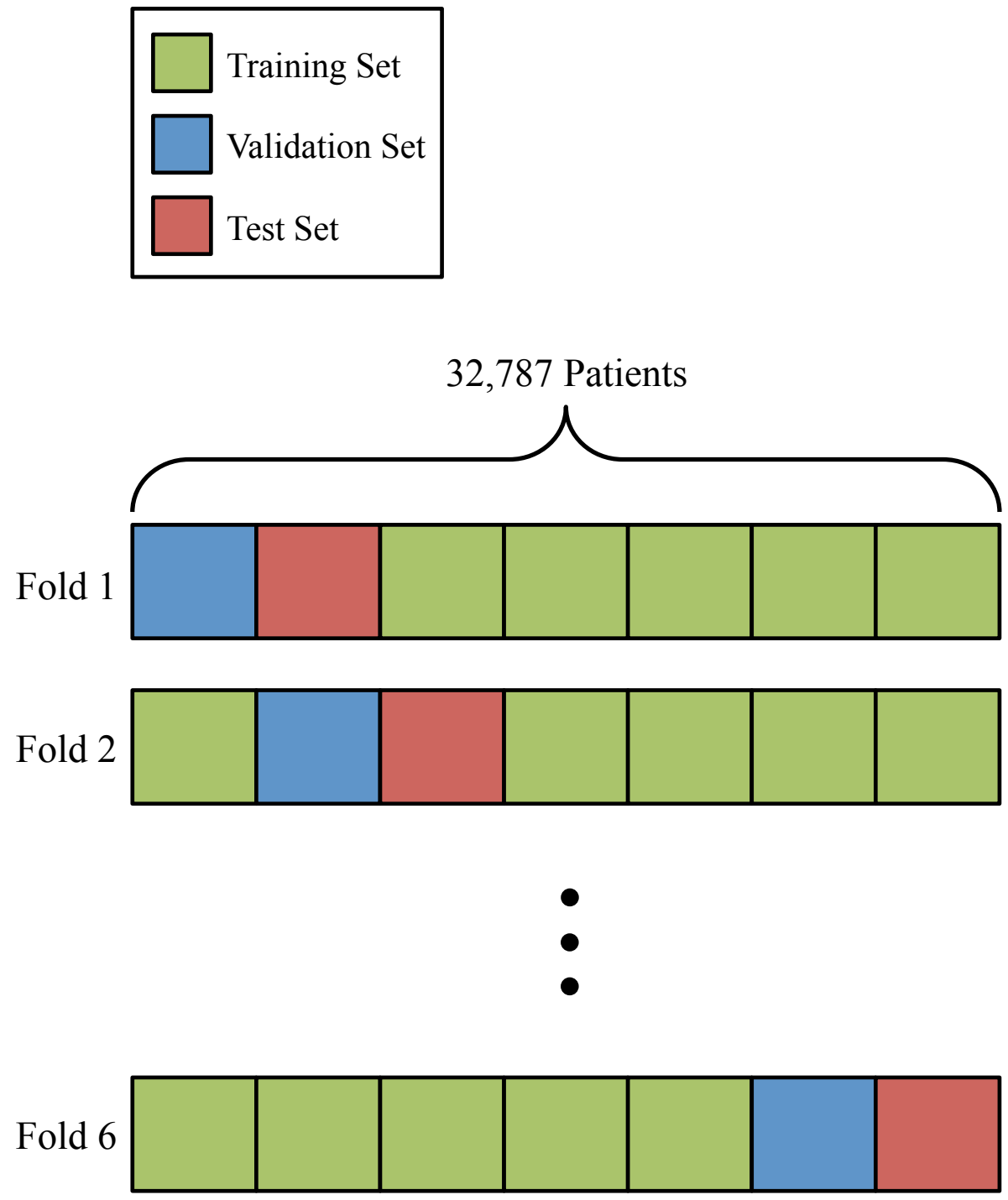

**Figure 9.** Diagram of 6-fold cross validation

Figure 6 depicts the 6-fold cross validation we performed for HF prediction. As explained earlier, the entire cohort is divided into 7 chunks, and two chunks take turn to play as the validation set and the test set.

**Hyper-parameters used for training models**

After experimenting with various values, the following hyper-parameter setting produced the best performance. We used Theano 0.7 and CUDA 7 for training logistic regression, MLP, and GRU models. SVM was implemented with Scikit-Learn Linear SVC. KNN was implemented with Scikit-Learn KNeighborsClassifier.

**Table 7.** Hyper-parameter settings for training the models

| Model | Hyper-parameter |
|---|---|
| Logistic regression, one-hot vectors | L2 regularization: 0.1, Max epoch: 100 |
| Logistic regression, medical concept vectors | L2 regularization: 0.01, Max epoch: 100 |
| SVM, one-hot vectors | L2 regularization: 0.000001, Dual: False |
| SVM, medical concept vectors | L2 regularization: 0.001, Dual: False |
| MLP, one-hot vectors | L2 regularization: 0.01, Hidden layer size: 15, Max epoch: 100 |
| MLP, medical concept vectors | L2 regularization: 0.001, Hidden layer size: 100, Max epoch: 100 |
| KNN, one-hot vectors | Number of neighbors: 15 |
| KNN, medical concept vectors | Number of neighbors: 100 |

## Complete table of 100 frequent diagnoses and their closest diagnoses, medications and procedures

| | Diagnoses | Medications | Procedures |
|---|---|---|---|
| Myalgia and myositis, unspecified (729.1) | -Lumbago (724.2)<br>-Thoracic or lumbosacral neuritis or radiculitis, unspecified (724.4)<br>-Degeneration of lumbar or lumbosacral intervertebral disc (722.52)<br>-Other malaise and fatigue (780.79) | -HYDROCODONE-ACETAMINOPHEN 5-500 MG PO TABS<br>-ZOLPIDEM TARTRATE 10 MG PO TABS<br>-CYCLOBENZAPRINE HCL 10 MG PO TABS<br>-HYDROCODONE-ACETAMINOPHEN 10-325 MG PO TABS<br>-TRAMADOL HCL 50 MG PO TABS | -PR INJ TRIGGER POINT(S) 1 OR 2 MUSCLES<br>-PHYSICAL MEDICINE, INT<br>-TRIAMCINOLONE ACETONIDE *KENALOG INJ<br>-ASP/INJ MAJOR JOINT/BURS/CYST<br>-PR METHLYPRED ACET (BU 21 - 40MG) *DEPO-MEDROL INJ |
| Pain in limb (729.5) | -Keratoderma, acquired (701.1)<br>-Plantar fascial fibromatosis (728.71)<br>-Other specified diseases of nail (703.8)<br>-Disturbance of skin sensation (782.0) | -HYDROCODONE-ACETAMINOPHEN 5-500 MG PO TABS | -DEBRIDEMENT OF NAILS, 6 OR MORE<br>-PODIATRY, INT<br>-REFERRAL TO PODIATRIC MEDICINE/SURGERY, INT<br>-PR OV EST PT LEV 3<br>-FOOT COMPLETE |
| Tear film insufficiency, unspecified (375.15) | -Senile cataract, unspecified (366.10)<br>-Presbyopia (367.4)<br>-Preglaucoma, unspecified (365.00)<br>-Other chronic allergic conjunctivitis (372.14) | -GLASSES<br>-ERYTHROMYCIN 5 MG/GM OP OINT<br>-PATANOL 0.1 % OP SOLN | -PR REFRACTION<br>-PR VISUAL FIELD EXAM EXTENDED<br>-PR VISUAL FIELD EXAM LIMITED<br>-REFERRAL TO OPHTHALMOLOGY, INT<br>-OPHTHALMOLOGY, INT |
| Headache (784.0) | -Dizziness and giddiness (780.4)<br>-Cervicalgia (723.1)<br>-Other malaise and fatigue (780.79)<br>-Acute sinusitis, unspecified (461.9) | -HYDROCODONE-ACETAMINOPHEN 5-500 MG PO TABS<br>-FLUTICASONE PROPIONATE 50 MCG/ACT NA SUSP<br>-CYCLOBENZAPRINE HCL 10 MG PO TABS<br>-FLONASE 50 MCG/ACT NA SUSP<br>-AZITHROMYCIN 250 MG PO TABS | -REFERRAL TO NEUROLOGY, INT<br>-NEUROLOGY, INT<br>-SERV PROV DURING REG SCHED EVE/WKEND/HOL HRS<br>-PR OV EST PT LEV 3<br>-PR ECG AND INTERPRETATION |
| Disturbance of skin sensation (782.0) | -Brachial neuritis or radiculitis NOS (723.4)<br>-Cervicalgia (723.1)<br>-Pain in limb (729.5)<br>-Headache (784.0) | -HYDROCODONE-ACETAMINOPHEN 5-500 MG PO TABS<br>-PREDNISONE 20 MG PO TABS | -NEUROLOGY, INT<br>-SENSE/MIXED N CONDUCTION TST<br>-REFERRAL TO NEUROLOGY, INT<br>-PR NERVE COND STUDY MOTOR W F-WAVE EA<br>-PR EMG NEEDLE ONE EXTREMITY |
| Edema (782.3) | -Congestive heart failure, unspecified (428.0)<br>-Unspecified essential hypertension (401.9)<br>-Atrial fibrillation (427.31)<br>-Chronic kidney disease, Stage III (moderate) (585.3) | -FUROSEMIDE 20 MG PO TABS<br>-HYDROCHLOROTHIAZIDE 25 MG PO TABS<br>-HYDROCODONE-ACETAMINOPHEN 5-500 MG PO TABS<br>-CEPHALEXIN 500 MG PO CAPS | -DEBRIDEMENT OF NAILS, 6 OR MORE<br>-PR OV EST PT MIN SERV<br>-EKG<br>-PR ECG AND INTERPRETATION<br>-CHEST PA & LATERAL |
| | | -FUROSEMIDE 40 MG PO TABS | |
| Asthma, unspecified type, unspecified (493.90) | -Allergic rhinitis, cause unspecified (477.9)<br>-Acute bronchitis (466.0)<br>-Bronchitis, not specified as acute or chronic (490)<br>-Acute upper respiratory infections of unspecified site (465.9) | -AZITHROMYCIN 250 MG PO TABS<br>-ALBUTEROL SULFATE HFA 108 (90 BASE) MCG/ACT IN AERS<br>-ALBUTEROL SULFATE HFA 108 MCG/ACT IN AERS<br>-PROMETHAZINE-CODEINE 6.25-10 MG/5ML PO SYRP<br>-ALBUTEROL 90 MCG/ACT IN AERS | -CHEST PA & LATERAL<br>-PULSE OXIMETRY SINGLE<br>-PR INHALATION RX FOR OBSTRUCTION MDI/NEB<br>-XR CHEST 2 VIEWS PA LATERAL<br>-IMMUN ADMIN IM/SQ/ID/PERC 1ST VAC ONLY |
| Routine gynecological examination (V72.31) | -Other screening mammogram (V76.12)<br>-Screening for lipoid disorders (V77.91)<br>-Routine general medical examination at a health care facility (V70.0)<br>-Screening for unspecified condition (V82.9) | -OBTAINING SCREEN PAP SMEAR<br>-GYN CYTOLOGY (PAP) PA<br>-MAMMOGRAM SCREENING<br>-MAMMO SCREENING BILATERAL DIGITAL<br>-SCRN MAMMO DIR DIG IMAGE BIL (V76.12) | |
| Diabetes mellitus without mention of complication, type II or unspecified type, uncontrolled (250.02) | -Mixed hyperlipidemia (272.2)<br>-Other abnormal glucose (790.29)<br>-Obesity, unspecified (278.00)<br>-Pure hypercholesterolemia (272.0) | -METFORMIN HCL 500 MG PO TABS<br>-METFORMIN HCL 1000 MG PO TABS<br>-GLUCOSE BLOOD VI STRP<br>-LISINOPRIL 10 MG PO TABS<br>-LISINOPRIL 20 MG PO TABS | -DIABETIC EYE EXAM (NO BILL)<br>-DIABETES EDUCATION, INT<br>-OPHTHALMOLOGY, INT<br>-DIABETIC FOOT EXAM (NO BILL)<br>-INFLUENZA VAC 3+YR (V04.81) IM |
| Need for prophylactic vaccination and inoculation against diphtheria-tetanus-pertussis, combined [DTP] [DTaP] (V06.1) | -Laboratory examination ordered as part of a routine general medical examination (V72.62)<br>-Need for prophylactic vaccination and inoculation against other combinations of diseases (V06.8)<br>-Screening for lipoid disorders (V77.91)<br>-Need for prophylactic vaccination and inoculation against influenza (V04.81) | -FLUTICASONE PROPIONATE 50 MCG/ACT NA SUSP<br>-AZITHROMYCIN 250 MG PO TABS<br>-ZOLPIDEM TARTRATE 10 MG PO TABS<br>-SIMVASTATIN 20 MG PO TABS | -TDAP VAC 11-64 YO *ADACEL (V06.8) IM<br>-MAMMO SCREENING BILATERAL DIGITAL<br>-PR INFLUENZA VACC PRES FREE 3+ YO 0.5ML IM<br>-REFERRAL TO GI, INT<br>-PNEUMOCOCCAL VAC (V03.82) IM/SQ |
| Diabetes mellitus without mention of complication, type II or unspecified type, not stated as uncontrolled (250.00) | -Unspecified essential hypertension (401.9)<br>-Diabetes mellitus without mention of complication, type II or unspecified type, uncontrolled (250.02)<br>-Obesity, unspecified (278.00)<br>-Benign essential hypertension (401.1) | -METFORMIN HCL 500 MG PO TABS<br>-METFORMIN HCL 1000 MG PO TABS<br>-GLUCOSE BLOOD VI STRP<br>-HYDROCHLOROTHIAZIDE 25 MG PO TABS<br>-LISINOPRIL 10 MG PO TABS | -DIABETIC EYE EXAM (NO BILL)<br>-PR DIABETIC EYE EXAM (NO BILL)<br>-INFLUENZA VAC 3+YR (V04.81) IM<br>-IMMUN ADMIN IM/SQ/ID/PERC 1ST VAC ONLY<br>-OPHTHALMOLOGY, INT |
| Personal history of colonic polyps (V12.72) | -Special screening for malignant neoplasms of colon (V76.51)<br>-Family history of malignant neoplasm of gastrointestinal tract (V16.0)<br>-Diverticulosis of colon (without mention of hemorrhage) (562.10)<br>-Routine general medical examination at a health care facility (V70.0) | -PEG-KCL-NACL-NASULF-NA ASC-C 100 G PO SOLR<br>-MOVIPREP 100 G PO SOLR<br>-SIMVASTATIN 20 MG PO TABS<br>-OMEPRAZOLE 20 MG PO CPDR | -PR COLONOSCOPY W/BIOPSY(S)<br>-PR COLONOSCOPY W/RMVL LESN/TUM/POLYP SNARE<br>-REFERRAL TO GI, INT<br>-GI, INT<br>-PR COLONOSCOPY DIAG |
| Acute upper respiratory infections of unspecified site (465.9) | -Cough (786.2)<br>-Acute sinusitis, unspecified (461.9)<br>-Acute bronchitis (466.0)<br>-Acute pharyngitis (462) | -AZITHROMYCIN 250 MG PO TABS<br>-PROMETHAZINE-CODEINE 6.25-10 MG/5ML PO SYRP<br>-AMOXICILLIN 500 MG PO CAPS<br>-FLUTICASONE PROPIONATE 50 MCG/ACT NA SUSP<br>-FLONASE 50 MCG/ACT NA SUSP | -PULSE OXIMETRY SINGLE<br>-SERV PROV DURING REG SCHED EVE/WKEND/HOL HRS<br>-CHEST PA & LATERAL<br>-GYN CYTOLOGY (PAP) PA<br>-INFLUENZA VAC (FLU CLINIC ONLY) 3+YO PA |
| Need for prophylactic vaccination and inoculation against tetanus-diphtheria [Td] (DT) (V06.5) | -Routine general medical examination at a health care facility (V70.0)<br>-Other general medical examination for administrative purposes (V70.3)<br>-Need for prophylactic vaccination and inoculation against other combinations of diseases (V06.8)<br>-Multiphasic screening (V82.6) | -TD VAC 7+ YO (V06.5) IM<br>-IMMUN ADMIN IM/SQ/ID/PERC 1ST VAC ONLY<br>-CHART ABSTRACTION SIMP (V70.3)<br>-MAMMOGRAM SCREENING<br>-IMMUN ADMIN, EACH ADD | |
| Esophageal reflux (530.81) | -Unspecified essential hypertension (401.9)<br>-Allergic rhinitis, cause unspecified (477.9)<br>-Cough (786.2)<br>-Routine general medical examination at a health care facility (V70.0) | -OMEPRAZOLE 20 MG PO CPDR<br>-ACIPHEX 20 MG PO TBEC<br>-PROTONIX 40 MG PO TBEC<br>-PRILOSEC 20 MG PO CPDR<br>-FLUTICASONE PROPIONATE 50 MCG/ACT NA SUSP | -GI, INT<br>-PR ECG AND INTERPRETATION<br>-PR LARYNGOSCOPY FLEX DIAG<br>-PR OV EST PT LEV 3<br>-CHEST PA & LATERAL |
| Other specified aftercare following surgery (V58.49) | -Lens replaced by other means (V43.1)<br>-Senile cataract, unspecified (366.10)<br>-Unspecified aftercare (V58.9)<br>-Follow-up examination, following surgery, unspecified (V67.00) | -GLASSES<br>-HYDROCODONE-ACETAMINOPHEN 5-500 MG PO TABS<br>-VICODIN 5-500 MG PO TABS<br>-CEPHALEXIN 500 MG PO CAPS<br>-IBUPROFEN 600 MG PO TABS | -PR ECG AND INTERPRETATION<br>-PR OPHTHALMIC BIOMETRY<br>-FOOT COMPLETE<br>-EKG<br>-PR REFRACTION |
| Diarrhea (787.91) | -Irritable bowel syndrome (564.1)<br>-Nausea with vomiting (787.01)<br>-Hemorrhage of rectum and anus (569.3)<br>-Nausea alone (787.02) | -METRONIDAZOLE 500 MG PO TABS<br>-CIPRO 500 MG PO TABS<br>-OMEPRAZOLE 20 MG PO CPDR<br>-CIPROFLOXACIN HCL 500 MG PO TABS<br>-PROMETHAZINE HCL 25 MG PO TABS | -GI, INT<br>-REFERRAL TO GI, INT<br>-SERV PROV DURING REG SCHED EVE/WKEND/HOL HRS<br>-PR COLONOSCOPY W/BIOPSY(S)<br>-US ABDOMEN COMPLETE |
| Coronary atherosclerosis of unspecified type of vessel, native or graft (414.00) | -Unspecified essential hypertension (401.9)<br>-Atrial fibrillation (427.31)<br>-Benign essential hypertension (401.1)<br>-Congestive heart failure, unspecified (428.0) | -SIMVASTATIN 40 MG PO TABS<br>-NITROGLYCERIN 0.4 MG SL SUBL<br>-SIMVASTATIN 20 MG PO TABS<br>-LISINOPRIL 10 MG PO TABS<br>-LISINOPRIL 20 MG PO TABS | -PR ECG AND INTERPRETATION<br>-EKG<br>-STRESS ECHO<br>-TTE W/O DOPPLER CMPLT<br>-PR CARDIAC STRESS COMPLTE |
| Dizziness and giddiness (780.4) | -Benign paroxysmal positional vertigo (386.11)<br>-Other malaise and fatigue (780.79)<br>-Unspecified hearing loss (389.9)<br>-Chest pain, unspecified (786.50) | -LORAZEPAM 0.5 MG PO TABS<br>-MECLIZINE HCL 25 MG PO TABS<br>-FLUTICASONE PROPIONATE 50 MCG/ACT NA SUSP | -PR ECG AND INTERPRETATION<br>-PR TYMPANOMETRY (IMPEDANCE TESTING)<br>-EKG<br>-PR COMPREHENSIVE HEARING TEST<br>-ENT, INT |
| Impacted cerumen (380.4) | -Unspecified hearing loss (389.9)<br>-Infective otitis externa, | -NEOMYCIN-POLYMYXIN-HC 1 % OT SOLN<br>-AMOXICILLIN 500 MG PO | -PR REMVL EAR WAX UNI/BILAT (380.4)<br>-ENT, INT |
| | unspecified (380.10)<br>-Dysfunction of Eustachian tube (381.81)<br>-Dizziness and giddiness (780.4) | CAPS<br>-FLONASE 50 MCG/ACT NA SUSP<br>-FLUTICASONE PROPIONATE 50 MCG/ACT NA SUSP | -PR COMPREHENSIVE HEARING TEST<br>-PR TYMPANOMETRY (IMPEDANCE TESTING)<br>-CERUMEN RMVL BY NURSE/MA (NO BILL) |
| Other chronic dermatitis due to solar radiation (692.74) | -Personal history of other malignant neoplasm of skin (V10.83)<br>-Other seborrheic keratosis (702.19)<br>-Benign neoplasm of other specified sites of skin (216.8)<br>-Inflamed seborrheic keratosis (702.11) | -TRIAMCINOLONE ACETONIDE 0.1 % EX CREA<br>-FLUOCINONIDE 0.05 % EX CREA | -PR DESTR PRE-MALIG LESN 1ST<br>-PR DESTR PRE-MALIG LESN 2-14<br>-PR BIOPSY OF SKIN/SQ/MUC MEMBR LESN SINGLE<br>-PR DESTRUCT BENIGN LESION UP TO 14 LESIONS (LIQUID NITROGEN)<br>-PR OV EST PT LEV 3 |
| Chronic kidney disease, Stage III (moderate) (585.3) | -Anemia, unspecified (285.9)<br>-Diabetes with renal manifestations, type II or unspecified type, not stated as uncontrolled (250.40)<br>-Proteinuria (791.0)<br>-Unspecified essential hypertension (401.9) | -AMLODIPINE BESYLATE 10 MG PO TABS<br>-FUROSEMIDE 20 MG PO TABS<br>-AMLODIPINE BESYLATE 5 MG PO TABS<br>-LISINOPRIL 40 MG PO TABS<br>-HYDROCHLOROTHIAZIDE 25 MG PO TABS | -PR PROTHROMBIN TIME<br>-PR TX/ PROPHY/ DX INJ SQ/ IM<br>-PR OV EST PT MIN SERV<br>-EKG<br>-DEBRIDEMENT OF NAILS, 6 OR MORE |
| Gout, unspecified (274.9) | -Unspecified essential hypertension (401.9)<br>-Other and unspecified hyperlipidemia (272.4)<br>-Long-term (current) use of other medications (V58.69)<br>-Screening for malignant neoplasms of prostate (V76.44) | -COLCHICINE 0.6 MG PO TABS<br>-ALLOPURINOL 100 MG PO TABS<br>-ALLOPURINOL 300 MG PO TABS<br>-INDOMETHACIN 50 MG PO CAPS<br>-ATENOLOL 50 MG PO TABS | -DEBRIDEMENT OF NAILS, 6 OR MORE<br>-TRIAMCINOLONE ACETONIDE *KENALOG INJ |
| Other specified diseases of nail (703.8) | -Keratoderma, acquired (701.1)<br>-Peripheral vascular disease, unspecified (443.9)<br>-Ingrowing nail (703.0)<br>-Diabetes with neurological manifestations, type II or unspecified type, not stated as uncontrolled (250.60) | -ECONAZOLE NITRATE 1 % EX CREA | -DEBRIDEMENT OF NAILS, 6 OR MORE<br>-TRIM SKIN LESIONS, 2 TO 4<br>-PARE CORN/CALLUS, SINGLE LESION<br>-PODIATRY, INT<br>-PR DESTR PRE-MALIG LESN 1ST |
| Preglaucoma, unspecified (365.00) | -Tear film insufficiency, unspecified (375.15)<br>-Glaucomatous atrophy [cupping] of optic disc (377.14)<br>-Vitreous degeneration (379.21)<br>-Lens replaced by other means (V43.1) | -GLASSES | -PR VISUAL FIELD EXAM EXTENDED<br>-PR FUNDAL PHOTOGRAPHY<br>-PR OPHTHALMIC DX IMAGING<br>-PR VISUAL FIELD EXAM LIMITED<br>-PR CMPTR OPHTH IMG OPTIC NERVE |
| Pre-operative examination, unspecified (V72.84) | -Pre-operative cardiovascular examination (V72.81)<br>-Other specified pre-operative examination (V72.83)<br>-Unspecified aftercare (V58.9)<br>-Senile cataract, unspecified (366.10) | -HYDROCODONE-ACETAMINOPHEN 5-500 MG PO TABS<br>-VICODIN 5-500 MG PO TABS<br>-IBUPROFEN 600 MG PO TABS | -PR ECG AND INTERPRETATION<br>-EKG<br>-CHEST PA & LATERAL<br>-PR OV EST PT LEV 3<br>-ECG AND INTERPRETATION (ORDERS ONLY) SZ |
| Cough (786.2) | -Acute upper respiratory infections of unspecified site (465.9)<br>-Bronchitis, not specified as acute or chronic (490) | -AZITHROMYCIN 250 MG PO TABS<br>-PROMETHAZINE-CODEINE 6.25-10 MG/5ML PO SYRP<br>-ALBUTEROL SULFATE HFA | -CHEST PA & LATERAL<br>-XR CHEST 2 VIEWS PA LATERAL<br>-PULSE OXIMETRY SINGLE<br>-SERV PROV DURING REG |
| | -Asthma, unspecified type, unspecified (493.90)<br>-Acute sinusitis, unspecified (461.9) | 108 (90 BASE) MCG/ACT IN AERS<br>-FLUTICASONE PROPIONATE 50 MCG/ACT NA SUSP<br>-GUAIFENESIN-CODEINE 100-10 MG/5ML PO SYRP | SCHED EVE/WKEND/HOL HRS<br>-PR ECG AND INTERPRETATION |
| Palpitations (785.1) | -Cardiac dysrhythmia, unspecified (427.9)<br>-Syncope and collapse (780.2)<br>-Dizziness and giddiness (780.4)<br>-Shortness of breath (786.05) | -ATENOLOL 25 MG PO TABS<br>-METOPROLOL SUCCINATE ER 25 MG PO TB24<br>-LORAZEPAM 0.5 MG PO TABS | -HOLTER 24 HR ECG<br>-PR ECG AND INTERPRETATION<br>-PR CARDIAC STRESS COMPLTE<br>-EKG<br>-TTE W/O DOPPLER CMPLT |
| Need for prophylactic vaccination and inoculation against influenza (V04.81) | -Other and unspecified hyperlipidemia (272.4)<br>-Need for prophylactic vaccination and inoculation against diphtheria-tetanus-pertussis, combined [DTP] [DTaP] (V06.1)<br>-Unspecified essential hypertension (401.9)<br>-Screening for unspecified condition (V82.9) | -AZITHROMYCIN 250 MG PO TABS<br>-FLUTICASONE PROPIONATE 50 MCG/ACT NA SUSP<br>-SIMVASTATIN 20 MG PO TABS<br>-HYDROCHLOROTHIAZIDE 25 MG PO TABS | -PR INFLUENZA VACC PRES FREE 3+ YO 0.5ML IM<br>-INFLUENZA VAC 3+YR (V04.81) IM<br>-INFLUENZA VAC (FLU CLINIC ONLY) 3+YO PA<br>-IMMUN ADMIN IM/SQ/ID/PERC 1ST VAC ONLY<br>-TDAP VAC 11-64 YO *ADACEL (V06.8) IM |
| Benign essential hypertension (401.1) | -Unspecified essential hypertension (401.9)<br>-Pure hypercholesterolemia (272.0)<br>-Mixed hyperlipidemia (272.2)<br>-Diabetes mellitus without mention of complication, type II or unspecified type, not stated as uncontrolled (250.00) | -HYDROCHLOROTHIAZIDE 25 MG PO TABS<br>-ATENOLOL 50 MG PO TABS<br>-LISINOPRIL 10 MG PO TABS<br>-LISINOPRIL 40 MG PO TABS<br>-LISINOPRIL 20 MG PO TABS | -PR ECG AND INTERPRETATION<br>-INFLUENZA VAC 3+YR (V04.81) IM<br>-IMMUN ADMIN IM/SQ/ID/PERC 1ST VAC ONLY<br>-GI, INT<br>-PR OV EST PT LEV 3 |
| Unspecified essential hypertension (401.9) | -Diabetes mellitus without mention of complication, type II or unspecified type, not stated as uncontrolled (250.00)<br>-Benign essential hypertension (401.1)<br>-Coronary atherosclerosis of unspecified type of vessel, native or graft (414.00)<br>-Obesity, unspecified (278.00) | -HYDROCHLOROTHIAZIDE 25 MG PO TABS<br>-LISINOPRIL 10 MG PO TABS<br>-ATENOLOL 50 MG PO TABS<br>-LISINOPRIL 20 MG PO TABS<br>-SIMVASTATIN 20 MG PO TABS | -PR ECG AND INTERPRETATION<br>-EKG<br>-INFLUENZA VAC 3+YR (V04.81) IM<br>-IMMUN ADMIN IM/SQ/ID/PERC 1ST VAC ONLY |
| Other screening mammogram (V76.12) | -Screening for malignant neoplasms of cervix (V76.2)<br>-Routine general medical examination at a health care facility (V70.0)<br>-Screening for unspecified condition (V82.9)<br>-Special screening for malignant neoplasms of colon (V76.51) | -FLUTICASONE PROPIONATE 50 MCG/ACT NA SUSP | -MAMMO SCREENING BILATERAL DIGITAL<br>-MAMMOGRAM SCREENING<br>-GYN CYTOLOGY (PAP) PA<br>-OBTAINING SCREEN PAP SMEAR<br>-PR OV EST PT LEV 3 |
| Elevated blood pressure reading without diagnosis of hypertension (796.2) | -Screening for unspecified condition (V82.9)<br>-Impaired fasting glucose (790.21)<br>-Family history of ischemic heart disease (V17.3)<br>-Obesity, unspecified (278.00) | -AZITHROMYCIN 250 MG PO TABS | -PR OV EST PT LEV 3<br>-PR OV EST PT LEV 2<br>-CHART ABSTRACTION SIMP (V70.3)<br>-PR ECG AND INTERPRETATION<br>-TD VAC 7+ YO (V06.5) IM |
| Symptomatic menopausal or female climacteric states (627.2) | -Other screening mammogram (V76.12)<br>-Routine gynecological examination (V72.31)<br>-Screening for malignant neoplasms of cervix (V76.2)<br>-Disorder of bone and cartilage, unspecified (733.90) | -ZOLPIDEM TARTRATE 10 MG PO TABS | -MAMMOGRAM SCREENING<br>-MAMMO SCREENING BILATERAL DIGITAL<br>-GYN CYTOLOGY (PAP) PA<br>-PR OV EST PT LEV 3<br>-OBTAINING SCREEN PAP SMEAR |
| Congestive heart failure, unspecified (428.0) | -Edema (782.3)<br>-Coronary atherosclerosis of unspecified type of vessel, native or graft (414.00)<br>-Aortic valve disorders (424.1)<br>-Mitral valve disorders (424.0) | -FUROSEMIDE 20 MG PO TABS<br>-FUROSEMIDE 40 MG PO TABS<br>-LASIX 20 MG PO TABS<br>-LASIX 40 MG PO TABS<br>-LISINOPRIL 5 MG PO TABS | -TTE W/O DOPPLER CMPLT<br>-PR OV EST PT MIN SERV<br>-TRANSTHORACIC COMPLETE, ECHO<br>-CHEST PA & LATERAL<br>-PR PROTHROMBIN TIME |
| Other and unspecified hyperlipidemia (272.4) | -Diabetes mellitus without mention of complication, type II or unspecified type, not stated as uncontrolled (250.00)<br>-Benign essential hypertension (401.1)<br>-Coronary atherosclerosis of unspecified type of vessel, native or graft (414.00)<br>-Need for prophylactic vaccination and inoculation against influenza (V04.81) | -SIMVASTATIN 20 MG PO TABS<br>-SIMVASTATIN 40 MG PO TABS<br>-HYDROCHLOROTHIAZIDE 25 MG PO TABS<br>-LISINOPRIL 10 MG PO TABS<br>-LIPITOR 10 MG PO TABS | -PR ECG AND INTERPRETATION<br>-IMMUN ADMIN IM/SQ/ID/PERC 1ST VAC ONLY<br>-PR OV EST PT LEV 3<br>-INFLUENZA VAC 3+YR (V04.81) IM<br>-PR INFLUENZA VACC PRES FREE 3+ YO 0.5ML IM |
| Pure hypercholesterolemia (272.0) | -Benign essential hypertension (401.1)<br>-Other and unspecified hyperlipidemia (272.4)<br>-Impaired fasting glucose (790.21)<br>-Diabetes mellitus without mention of complication, type II or unspecified type, uncontrolled (250.02) | -SIMVASTATIN 20 MG PO TABS<br>-SIMVASTATIN 40 MG PO TABS<br>-LISINOPRIL 10 MG PO TABS<br>-ATENOLOL 50 MG PO TABS<br>-HYDROCHLOROTHIAZIDE 25 MG PO TABS | -EKG<br>-PR ECG AND INTERPRETATION<br>-TDAP VAC 11-64 YO *ADACEL (V06.8) IM<br>-MAMMO SCREENING BILATERAL DIGITAL |
| Mixed hyperlipidemia (272.2) | -Benign essential hypertension (401.1)<br>-Diabetes mellitus without mention of complication, type II or unspecified type, uncontrolled (250.02)<br>-Impaired fasting glucose (790.21)<br>-Obesity, unspecified (278.00) | -SIMVASTATIN 40 MG PO TABS<br>-SIMVASTATIN 20 MG PO TABS<br>-LISINOPRIL 10 MG PO TABS<br>-SIMVASTATIN 10 MG PO TABS<br>-METFORMIN HCL 500 MG PO TABS | -TDAP VAC 11-64 YO *ADACEL (V06.8) IM<br>-INFLUENZA VAC 3+YR (V04.81) IM<br>-EKG<br>-PR CARDIAC STRESS COMPLTE |
| Unspecified acquired hypothyroidism (244.9) | -Unspecified essential hypertension (401.9)<br>-Benign essential hypertension (401.1)<br>-Other screening mammogram (V76.12)<br>-Other malaise and fatigue (780.79) | -LEVOTHYROXINE SODIUM 75 MCG PO TABS<br>-HYDROCHLOROTHIAZIDE 25 MG PO TABS<br>-LEVOTHYROXINE SODIUM 100 MCG PO TABS<br>-SIMVASTATIN 20 MG PO TABS<br>-SIMVASTATIN 40 MG PO TABS | -MAMMO SCREENING BILATERAL DIGITAL<br>-IMMUN ADMIN IM/SQ/ID/PERC 1ST VAC ONLY<br>-PR OV EST PT LEV 3<br>-MAMMOGRAM SCREENING<br>-PR INFLUENZA VACC PRES FREE 3+ YO 0.5ML IM |
| Depressive disorder, not elsewhere classified (311) | -Insomnia, unspecified (780.52)<br>-Other malaise and fatigue (780.79)<br>-Unspecified essential hypertension (401.9) | -ZOLPIDEM TARTRATE 10 MG PO TABS<br>-LORAZEPAM 0.5 MG PO TABS<br>-HYDROCHLOROTHIAZIDE 25 MG PO TABS<br>-TRAZODONE HCL 50 MG PO | -IMMUN ADMIN IM/SQ/ID/PERC 1ST VAC ONLY<br>-PR OV EST PT LEV 3<br>-INFLUENZA VAC 3+YR (V04.81) IM |
| | -Other and unspecified hyperlipidemia (272.4) | TABS<br>-HYDROCODONE-ACETAMINOPHEN 5-500 MG PO TABS | -MAMMO SCREENING BILATERAL DIGITAL |
| Personal history of other malignant neoplasm of skin (V10.83) | -Other chronic dermatitis due to solar radiation (692.74)<br>-Benign neoplasm of other specified sites of skin (216.8)<br>-Other seborrheic keratosis (702.19)<br>-Neoplasm of uncertain behavior of skin (238.2) | -FLUOCINONIDE 0.05 % EX CREA<br>-TRIAMCINOLONE ACETONIDE 0.1 % EX CREA | -PR DESTR PRE-MALIG LESN 1ST<br>-PR DESTR PRE-MALIG LESN 2-14<br>-PR BIOPSY OF SKIN/SQ/MUC MEMBR LESN SINGLE<br>-SURGICAL PATHOLOGY PA<br>-PR OV EST PT LEV 3 |
| Bronchitis, not specified as acute or chronic (490) | -Acute upper respiratory infections of unspecified site (465.9)<br>-Cough (786.2)<br>-Acute sinusitis, unspecified (461.9)<br>-Acute nasopharyngitis [common cold] (460) | -PROMETHAZINE-CODEINE 6.25-10 MG/5ML PO SYRP<br>-AZITHROMYCIN 250 MG PO TABS<br>-ZITHROMAX Z-PAK 250 MG PO TABS<br>-ALBUTEROL 90 MCG/ACT IN AERS<br>-ALBUTEROL SULFATE HFA 108 MCG/ACT IN AERS | -PULSE OXIMETRY SINGLE<br>-CHEST PA & LATERAL<br>-PR INHALATION RX FOR OBSTRUCTION MDI/NEB<br>-SERV PROV DURING REG SCHED EVE/WKEND/HOL HRS<br>-XR CHEST 2 VIEWS PA LATERAL |
| Obstructive sleep apnea (adult)(pediatric) (327.23) | -Sleep disturbance, unspecified (780.50)<br>-Obesity, unspecified (278.00)<br>-Other respiratory abnormalities (786.09)<br>-Other malaise and fatigue (780.79) | -ZOLPIDEM TARTRATE 10 MG PO TABS<br>-FLUTICASONE PROPIONATE 50 MCG/ACT NA SUSP<br>-SIMVASTATIN 20 MG PO TABS<br>-ZOLPIDEM TARTRATE 5 MG PO TABS<br>-AZITHROMYCIN 250 MG PO TABS | -REFERRAL TO SLEEP DISORDERS, INT<br>-PR POLYSOMNOGRAPHY 4 OR MORE<br>-PR POLYSOMNOGRAPHY W/C P<br>-SLEEP STUDY ATTENDED<br>-CPAP EQUIPMENT DME |
| Anemia, unspecified (285.9) | -Chronic kidney disease, Stage III (moderate) (585.3)<br>-Edema (782.3)<br>-Unspecified essential hypertension (401.9)<br>-Iron deficiency anemia, unspecified (280.9) | -HYDROCHLOROTHIAZIDE 25 MG PO TABS | -INFLUENZA VAC 3+YR (V04.81) IM<br>-PR ECG AND INTERPRETATION<br>-GI, INT<br>-EKG<br>-CHEST PA & LATERAL |
| Lens replaced by other means (V43.1) | -Tear film insufficiency, unspecified (375.15)<br>-Vitreous degeneration (379.21)<br>-Regular astigmatism (367.21)<br>-Other specified aftercare following surgery (V58.49) | -GLASSES | -PR VISUAL FIELD EXAM EXTENDED<br>-PR REFRACTION<br>-PR OPHTHALMIC DX IMAGING<br>-DIABETIC EYE EXAM (NO BILL)<br>-PR DIABETIC EYE EXAM (NO BILL) |
| Chest pain, unspecified (786.50) | -Palpitations (785.1)<br>-Other chest pain (786.59)<br>-Other respiratory abnormalities (786.09)<br>-Other malaise and fatigue (780.79) | -AZITHROMYCIN 250 MG PO TABS<br>-NITROGLYCERIN 0.4 MG SL SUBL<br>-HYDROCODONE-ACETAMINOPHEN 5-500 MG PO TABS<br>-OMEPRAZOLE 20 MG PO CPDR | -PR ECG AND INTERPRETATION<br>-PR CARDIAC STRESS COMPLTE<br>-EKG<br>-STRESS ECHO<br>-CHEST PA & LATERAL |
| Acute sinusitis, unspecified (461.9) | -Cough (786.2)<br>-Bronchitis, not specified as acute or chronic (490)<br>-Allergic rhinitis, cause unspecified (477.9)<br>-Acute pharyngitis (462) | -AMOXICILLIN 500 MG PO CAPS<br>-AZITHROMYCIN 250 MG PO TABS<br>-FLONASE 50 MCG/ACT NA SUSP<br>-PROMETHAZINE-CODEINE 6.25-10 MG/5ML PO SYRP | -PULSE OXIMETRY SINGLE<br>-SERV PROV DURING REG SCHED EVE/WKEND/HOL HRS<br>-CHEST PA & LATERAL<br>-IMMUN ADMIN IM/SQ/ID/PERC 1ST VAC |
| | | -FLUTICASONE PROPIONATE 50 MCG/ACT NA SUSP | ONLY<br>-ENT, INT |
| Atrial fibrillation (427.31) | -Encounter for therapeutic drug monitoring (V58.83)<br>-Coronary atherosclerosis of unspecified type of vessel, native or graft (414.00)<br>-Congestive heart failure, unspecified (428.0)<br>-Unspecified essential hypertension (401.9) | -WARFARIN SODIUM 5 MG PO TABS<br>-ATENOLOL 50 MG PO TABS<br>-ATENOLOL 25 MG PO TABS<br>-METOPROLOL SUCCINATE ER 50 MG PO TB24<br>-SIMVASTATIN 40 MG PO TABS | -PR OV EST PT MIN SERV<br>-PR PROTHROMBIN TIME<br>-EKG<br>-PR ECG AND INTERPRETATION<br>-TTE W/O DOPPLER CMPLT |
| Urinary tract infection, site not specified (599.0) | -Dysuria (788.1)<br>-Urinary frequency (788.41)<br>-Other malaise and fatigue (780.79)<br>-Vaginitis and vulvovaginitis, unspecified (616.10) | -NITROFURANTOIN MONOHYD MACRO 100 MG PO CAPS<br>-CIPROFLOXACIN HCL 250 MG PO TABS<br>-SULFAMETHOXAZOLE-TMP DS 800-160 MG PO TABS<br>-MACROBID 100 MG PO CAPS<br>-CIPRO 500 MG PO TABS | -SERV PROV DURING REG SCHED EVE/WKEND/HOL HRS<br>-PR URINALYSIS AUTO W/O SCOPE<br>-PR ECG AND INTERPRETATION<br>-CYSTOURETHROSCOPY<br>-PR URINALYSIS DIP W/O SCOPE |
| Pain in joint, shoulder region (719.41) | -Other affections of shoulder region, not elsewhere classified (726.2)<br>-Rotator cuff (capsule) sprain (840.4)<br>-Cervicalgia (723.1)<br>-Pain in joint, lower leg (719.46) | -HYDROCODONE-ACETAMINOPHEN 5-500 MG PO TABS<br>-VICODIN 5-500 MG PO TABS<br>-NAPROXEN 500 MG PO TABS | -SHOULDER 2+ VIEWS<br>-ASP/INJ MAJOR JOINT/BURS/CYST<br>-TRIAMCINOLONE ACETONIDE *KENALOG INJ<br>-UPPER EXTREM JOINT MRI<br>-PR THERAPEUTIC EXERCISES |
| Encounter for therapeutic drug monitoring (V58.83) | -Atrial fibrillation (427.31)<br>-Congestive heart failure, unspecified (428.0)<br>-Coronary atherosclerosis of unspecified type of vessel, native or graft (414.00) | -WARFARIN SODIUM 5 MG PO TABS<br>-COUMADIN 5 MG PO TABS<br>-WARFARIN SODIUM 2.5 MG PO TABS<br>-ATENOLOL 50 MG PO TABS<br>-WARFARIN SODIUM 2 MG PO TABS | -PR OV EST PT MIN SERV<br>-PR PROTHROMBIN TIME<br>-DEBRIDEMENT OF NAILS, 6 OR MORE<br>-PR DESTR PRE-MALIG LESN 2-14<br>-PR ECG AND INTERPRETATION |
| Malignant neoplasm of breast (female), unspecified (174.9) | -Lump or mass in breast (611.72)<br>-Unspecified aftercare (V58.9)<br>-Disorder of bone and cartilage, unspecified (733.90)<br>-Pre-operative examination, unspecified (V72.84) | -LORAZEPAM 1 MG PO TABS<br>-LORAZEPAM 0.5 MG PO TABS<br>-TAMOXIFEN CITRATE 20 MG PO TABS<br>-VICODIN 5-500 MG PO TABS<br>-ANASTROZOLE 1 MG PO TABS | -MAMMOGRAM BILAT DIAGNOSTIC<br>-DEXA BONE DENSITY AXIAL SKELETON HIPS/PELV/SPINE<br>-MAMMO DIAGNOSTIC BILATERAL DIGITAL<br>-VENIPUNC FNGR,HEEL,EAR<br>-REFERRAL TO ONCOLOGY, INT |
| Contact dermatitis and other eczema, unspecified cause (692.9) | -Unspecified disorder of skin and subcutaneous tissue (709.9)<br>-Other atopic dermatitis and related conditions (691.8)<br>-Other seborrheic keratosis (702.19)<br>-Other chronic dermatitis due to solar radiation (692.74) | -TRIAMCINOLONE ACETONIDE 0.1 % EX CREA<br>-FLUOCINONIDE 0.05 % EX CREA<br>-TRIAMCINOLONE ACETONIDE 0.1 % EX OINT<br>-DESONIDE 0.05 % EX CREA<br>-FLUTICASONE PROPIONATE 50 MCG/ACT NA SUSP | -DERMATOLOGY, INT<br>-PR BIOPSY OF SKIN/SQ/MUC MEMBR LESN SINGLE<br>-PR DESTR PRE-MALIG LESN 1ST<br>-REFERRAL TO DERMATOLOGY, INT<br>-PR OV EST PT LEV 3 |
| Benign neoplasm of colon (211.3) | -Special screening for malignant neoplasms of colon (V76.51)<br>-Diverticulosis of colon (without mention of | -PEG-KCL-NACL-NASULF-NA ASC-C 100 G PO SOLR<br>-MOVIPREP 100 G PO SOLR<br>-OMEPRAZOLE 20 MG PO CPDR | -PR COLONOSCOPY W/BIOPSY(S)<br>-PR COLONOSCOPY W/RMVL LESN/TUM/POLYP SNARE |
| | hemorrhage) (562.10)<br>-Family history of malignant neoplasm of gastrointestinal tract (V16.0)<br>-Hemorrhage of rectum and anus (569.3) | -SIMVASTATIN 20 MG PO TABS | -REFERRAL TO GI, INT<br>-GI, INT<br>-PR COLONOSCOPY DIAG |
| Disorder of bone and cartilage, unspecified (733.90) | -Other screening mammogram (V76.12)<br>-Symptomatic menopausal or female climacteric states (627.2)<br>-Routine general medical examination at a health care facility (V70.0)<br>-Special screening for osteoporosis (V82.81) | -FOSAMAX 70 MG PO TABS | -DEXA BONE DENSITY AXIAL SKELETON HIPS/PELV/SPINE<br>-BONE DENSITY (DEXA)<br>-MAMMO SCREENING BILATERAL DIGITAL<br>-DEXA BONE DENSITY STUDY 1+ SITES AXIAL SKEL (HIP/PELVIS/SPINE)<br>-PR OV EST PT LEV 3 |
| Presbyopia (367.4) | -Myopia (367.1)<br>-Hypermetropia (367.0)<br>-Astigmatism, unspecified (367.20)<br>-Other specified visual disturbances (368.8) | -GLASSES | -PR REFRACTION<br>-DIABETIC EYE EXAM (NO BILL)<br>-PR VISUAL FIELD EXAM EXTENDED<br>-INFLUENZA VAC (FLU CLINIC ONLY) 3+YO PA<br>-PR VISUAL FIELD EXAM LIMITED |
| Screening for lipoid disorders (V77.91) | -Screening for malignant neoplasms of cervix (V76.2)<br>-Routine general medical examination at a health care facility (V70.0)<br>-Routine gynecological examination (V72.31)<br>-Screening for unspecified condition (V82.9) | -CHART ABSTRACTION SIMP (V70.3)<br>-PR OV EST PT LEV 2<br>-TDAP VAC 11-64 YO *ADACEL (V06.8) IM<br>-GYN CYTOLOGY (PAP) PA<br>-PR OV EST PT LEV 3 | |
| Myopia (367.1) | -Regular astigmatism (367.21)<br>-Astigmatism, unspecified (367.20)<br>-Other specified visual disturbances (368.8)<br>-Hypermetropia (367.0) | -GLASSES | -PR REFRACTION<br>-CONTACT LENS MISC PA<br>-INFLUENZA VAC (FLU CLINIC ONLY) 3+YO PA<br>-PR VISUAL FIELD EXAM EXTENDED<br>-DIABETIC EYE EXAM (NO BILL) |
| Acute pharyngitis (462) | -Acute sinusitis, unspecified (461.9)<br>-Acute nasopharyngitis [common cold] (460)<br>-Bronchitis, not specified as acute or chronic (490)<br>-Conjunctivitis, unspecified (372.30) | -AZITHROMYCIN 250 MG PO TABS<br>-AMOXICILLIN 500 MG PO CAPS<br>-FLONASE 50 MCG/ACT NA SUSP<br>-PROMETHAZINE-CODEINE 6.25-10 MG/5ML PO SYRP<br>-AUGMENTIN 875-125 MG PO TABS | -PR STREP A RAPID ASSAY W/OPTIC<br>-SERV PROV DURING REG SCHED EVE/WKEND/HOL HRS<br>-PULSE OXIMETRY SINGLE<br>-ENT, INT<br>-CHART ABSTRACTION SIMP (V70.3) |
| Screening for unspecified condition (V82.9) | -Other screening mammogram (V76.12)<br>-Other general medical examination for administrative purposes (V70.3)<br>-Need for prophylactic vaccination and inoculation against tetanus-diphtheria [Td] (DT) (V06.5)<br>-Screening for malignant neoplasms of cervix (V76.2) | -PR OV EST PT LEV 3<br>-PR OV EST PT LEV 2<br>-IMMUN ADMIN IM/SQ/ID/PERC 1ST VAC ONLY<br>-MAMMOGRAM SCREENING<br>-GYN CYTOLOGY (PAP) PA | |
| Screening for malignant neoplasms of cervix (V76.2) | -Other screening mammogram (V76.12)<br>-Screening for lipoid disorders (V77.91)<br>-Routine general medical examination at a health care facility (V70.0)<br>-Screening for unspecified condition (V82.9) | -GYN CYTOLOGY (PAP) PA<br>-OBTAINING SCREEN PAP SMEAR<br>-MAMMOGRAM SCREENING<br>-PR OV EST PT LEV 3<br>-PR OV EST PT LEV 2 | |
| Impotence of organic origin (607.84) | -Hypertrophy (benign) of prostate without urinary obstruction and other lower urinary tract symptom (LUTS) (600.00)<br>-Other and unspecified hyperlipidemia (272.4)<br>-Hypertrophy (benign) of prostate with urinary obstruction and other lower urinary tract symptoms (LUTS) (600.01)<br>-Unspecified essential hypertension (401.9) | -VIAGRA 100 MG PO TABS<br>-SILDENAFIL CITRATE 100 MG PO TABS<br>-VIAGRA 50 MG PO TABS<br>-TADALAFIL 20 MG PO TABS<br>-SIMVASTATIN 20 MG PO TABS | -PR OV EST PT LEV 3<br>-IMMUN ADMIN IM/SQ/ID/PERC 1ST VAC ONLY<br>-PR DESTR PRE-MALIG LESN 1ST<br>-PR OV EST PT LEV 2<br>-INFLUENZA VAC 3+YR (V04.81) IM |
| Unspecified sinusitis (chronic) (473.9) | -Acute sinusitis, unspecified (461.9)<br>-Allergic rhinitis, cause unspecified (477.9)<br>-Dysfunction of Eustachian tube (381.81)<br>-Headache (784.0) | -AMOXICILLIN-POT CLAVULANATE 875-125 MG PO TABS<br>-FLUTICASONE PROPIONATE 50 MCG/ACT NA SUSP<br>-MOMETASONE FUROATE 50 MCG/ACT NA SUSP<br>-FLONASE 50 MCG/ACT NA SUSP<br>-AUGMENTIN 875-125 MG PO TABS | -PR NASAL ENDO DIAG<br>-ENT, INT<br>-CT MAXILLOFACIAL W/O CONTRAST<br>-PR LARYNGOSCOPY FLEX DIAG<br>-REFERRAL TO ENT, INT |
| Need for prophylactic vaccination and inoculation against viral hepatitis (V05.3) | -Other specified counseling (V65.49)<br>-Need for prophylactic vaccination and inoculation against other combinations of diseases (V06.8)<br>-Need for prophylactic vaccination and inoculation against poliomyelitis (V04.0)<br>-Need for prophylactic vaccination and inoculation against tetanus-diphtheria [Td] (DT) (V06.5) | -CIPROFLOXACIN HCL 500 MG PO TABS<br>-MEFLOQUINE HCL 250 MG PO TABS | -HEP A VAC ADULT (V05.3) IM<br>-IMMUN ADMIN, EACH ADD<br>-HEP B VAC ADULT (V05.3) IM<br>-IMMUN ADMIN IM/SQ/ID/PERC 1ST VAC ONLY<br>-TDAP VAC 11-64 YO *ADACEL (V06.8) IM |
| Senile cataract, unspecified (366.10) | -Presbyopia (367.4)<br>-Regular astigmatism (367.21)<br>-Preglaucoma, unspecified (365.00)<br>-Tear film insufficiency, unspecified (375.15) | -GLASSES | -PR REFRACTION<br>-PR VISUAL FIELD EXAM EXTENDED<br>-DIABETIC EYE EXAM (NO BILL)<br>-PR VISUAL FIELD EXAM LIMITED<br>-PR OPHTHALMIC DX IMAGING |
| Insomnia, unspecified (780.52) | -Depressive disorder, not elsewhere classified (311)<br>-Other malaise and fatigue (780.79)<br>-Sleep disturbance, unspecified (780.50)<br>-Need for prophylactic vaccination and inoculation against influenza (V04.81) | -ZOLPIDEM TARTRATE 10 MG PO TABS<br>-AMBIEN 10 MG PO TABS<br>-LORAZEPAM 0.5 MG PO TABS<br>-ZOLPIDEM TARTRATE 5 MG PO TABS<br>-TRAZODONE HCL 50 MG PO TABS | -PR INFLUENZA VACC PRES FREE 3+ YO 0.5ML IM<br>-MAMMO SCREENING BILATERAL DIGITAL<br>-PR OV EST PT LEV 3<br>-IMMUN ADMIN IM/SQ/ID/PERC 1ST VAC ONLY<br>-TDAP VAC 11-64 YO *ADACEL (V06.8) IM |
| Osteoarthrosis, localized, primary, lower leg (715.16) | -Osteoarthrosis, localized, not specified whether | -HYDROCODONE-ACETAMINOPHEN 10-325 | -ASP/INJ MAJOR JOINT/BURS/CYST |
| | primary or secondary, lower leg (715.36)<br>-Chondromalacia of patella (717.7)<br>-Pain in joint, pelvic region and thigh (719.45)<br>-Enthesopathy of hip region (726.5) | MG PO TABS<br>-HYDROCODONE-ACETAMINOPHEN 5-500 MG PO TABS | -TRIAMCINOLONE ACETONIDE *KENALOG INJ<br>-REFERRAL TO ORTHOPEDICS, INT<br>-KNEE 3 VIEW<br>-BETAMETH ACET 3MG & NA PHOS 3MG *CELESTONE INJ |
| Obesity, unspecified (278.00) | -Routine general medical examination at a health care facility (V70.0)<br>-Other and unspecified hyperlipidemia (272.4)<br>-Unspecified sleep apnea (780.57)<br>-Unspecified essential hypertension (401.9) | -METFORMIN HCL 500 MG PO TABS<br>-HYDROCHLOROTHIAZIDE 25 MG PO TABS | -PR OV EST PT LEV 3<br>-NUTRITION, INT<br>-PR ECG AND INTERPRETATION<br>-IMMUN ADMIN IM/SQ/ID/PERC 1ST VAC ONLY<br>-PR OV EST PT LEV 2 |
| Screening for malignant neoplasms of prostate (V76.44) | -Special screening for malignant neoplasms of colon (V76.51)<br>-Screening for unspecified condition (V82.9)<br>-Impaired fasting glucose (790.21)<br>-Impotence of organic origin (607.84) | -SIMVASTATIN 20 MG PO TABS | -IMMUN ADMIN IM/SQ/ID/PERC 1ST VAC ONLY<br>-PR OV EST PT LEV 3<br>-PR OV EST PT LEV 2<br>-GI, INT<br>-TDAP VAC 11-64 YO *ADACEL (V06.8) IM |
| Unspecified vitamin D deficiency (268.9) | -Laboratory examination ordered as part of a routine general medical examination (V72.62)<br>-Other malaise and fatigue (780.79)<br>-Need for prophylactic vaccination and inoculation against diphtheria-tetanus-pertussis, combined [DTP] [DTaP] (V06.1)<br>-Anemia, unspecified (285.9) | -ERGOCALCIFEROL 50000 UNITS PO CAPS<br>-FLUTICASONE PROPIONATE 50 MCG/ACT NA SUSP<br>-SIMVASTATIN 20 MG PO TABS<br>-OMEPRAZOLE 20 MG PO CPDR<br>-ZOLPIDEM TARTRATE 10 MG PO TABS | -MAMMO SCREENING BILATERAL DIGITAL<br>-TDAP VAC 11-64 YO *ADACEL (V06.8) IM<br>-DEXA BONE DENSITY AXIAL SKELETON HIPS/PELV/SPINE<br>-EKG<br>-REFERRAL TO GI, INT |
| Lumbago (724.2) | -Sciatica (724.3)<br>-Cervicalgia (723.1)<br>-Thoracic or lumbosacral neuritis or radiculitis, unspecified (724.4)<br>-Degeneration of lumbar or lumbosacral intervertebral disc (722.52) | -HYDROCODONE-ACETAMINOPHEN 5-500 MG PO TABS<br>-CYCLOBENZAPRINE HCL 10 MG PO TABS<br>-FLEXERIL 10 MG PO TABS<br>-VICODIN 5-500 MG PO TABS<br>-CARISOPRODOL 350 MG PO TABS | -LS SPINE 2 VIEWS<br>-LUMBAR SPINE MRI W/O CONTR<br>-PHYSICAL MEDICINE, INT<br>-PHYSICAL THERAPY, INT<br>-PHYSICAL MEDICINE, INT |
| Backache, unspecified (724.5) | -Sciatica (724.3)<br>-Cervicalgia (723.1)<br>-Thoracic or lumbosacral neuritis or radiculitis, unspecified (724.4)<br>-Pain in joint, pelvic region and thigh (719.45) | -HYDROCODONE-ACETAMINOPHEN 5-500 MG PO TABS<br>-CYCLOBENZAPRINE HCL 10 MG PO TABS<br>-NAPROXEN 500 MG PO TABS<br>-FLEXERIL 10 MG PO TABS<br>-VICODIN 5-500 MG PO TABS | -LS SPINE 2 VIEWS<br>-PHYSICAL MEDICINE, INT<br>-PR ECG AND INTERPRETATION<br>-LUMBAR SPINE MRI W/O CONTR<br>-PR OV EST PT LEV 3 |
| Thoracic or lumbosacral neuritis or radiculitis, unspecified (724.4) | -Spinal stenosis, lumbar region, without neurogenic claudication (724.02)<br>-Displacement of lumbar intervertebral disc without myelopathy (722.10)<br>-Sciatica (724.3)<br>-Lumbago (724.2) | -HYDROCODONE-ACETAMINOPHEN 10-325 MG PO TABS<br>-HYDROCODONE-ACETAMINOPHEN 5-500 MG PO TABS<br>-GABAPENTIN 300 MG PO CAPS<br>-CYCLOBENZAPRINE HCL 10 MG PO TABS<br>-PREDNISONE 20 MG PO TABS | -LUMBAR SPINE MRI W/O CONTR<br>-MRI LUMBAR SPINE WO CONTRAST<br>-REFERRAL TO PHYSICAL MEDICINE, INT<br>-PHYSICAL MEDICINE, INT<br>-PHYSICAL MEDICINE, INT |
| Inflamed seborrheic keratosis (702.11) | -Other chronic dermatitis due to solar radiation (692.74)<br>-Other dyschromia (709.09)<br>-Viral warts, unspecified (078.10)<br>-Nevus, non-neoplastic (448.1) | -TRIAMCINOLONE ACETONIDE 0.1 % EX CREA<br>-ECONAZOLE NITRATE 1 % EX CREA | -PR DESTRUCT BENIGN LESION UP TO 14 LESIONS (LIQUID NITROGEN)<br>-PR DESTR PRE-MALIG LESN 1ST<br>-PR DESTR PRE-MALIG LESN 2-14<br>-PR BIOPSY OF SKIN/SQ/MUC MEMBR LESN SINGLE<br>-DERMATOLOGY, INT |
| Other seborrheic keratosis (702.19) | -Actinic keratosis (702.0)<br>-Inflamed seborrheic keratosis (702.11)<br>-Other dyschromia (709.09)<br>-Personal history of other malignant neoplasm of skin (V10.83) | -TRIAMCINOLONE ACETONIDE 0.1 % EX CREA<br>-FLUOCINONIDE 0.05 % EX CREA | -PR DESTR PRE-MALIG LESN 1ST<br>-PR DESTR PRE-MALIG LESN 2-14<br>-PR BIOPSY OF SKIN/SQ/MUC MEMBR LESN SINGLE<br>-PR DESTRUCT BENIGN LESION UP TO 14 LESIONS (LIQUID NITROGEN)<br>-PR OV EST PT LEV 3 |
| Allergic rhinitis due to other allergen (477.8) | -Allergic rhinitis due to pollen (477.0)<br>-Asthma, unspecified type, unspecified (493.90)<br>-Extrinsic asthma, unspecified (493.00)<br>-Acute sinusitis, unspecified (461.9) | -FLUTICASONE PROPIONATE 50 MCG/ACT NA SUSP<br>-FLONASE 50 MCG/ACT NA SUSP<br>-MOMETASONE FUROATE 50 MCG/ACT NA SUSP<br>-ALLEGRA 180 MG PO TABS<br>-NASONEX 50 MCG/ACT NA SUSP | -IMMUNOTHERAPY MULTIPLE<br>-ANTIGEN ALLERGY MAT FOR INJ (06041) PA<br>-PR IMMUNOTX W EXTRACT 2+ INJ<br>-INFLUENZA VAC (FLU CLINIC ONLY) 3+YO PA<br>-ENT, INT |
| Allergic rhinitis, cause unspecified (477.9) | -Acute sinusitis, unspecified (461.9)<br>-Routine general medical examination at a health care facility (V70.0)<br>-Acute upper respiratory infections of unspecified site (465.9)<br>-Cough (786.2) | -FLUTICASONE PROPIONATE 50 MCG/ACT NA SUSP<br>-FLONASE 50 MCG/ACT NA SUSP<br>-MOMETASONE FUROATE 50 MCG/ACT NA SUSP<br>-AZITHROMYCIN 250 MG PO TABS<br>-NASONEX 50 MCG/ACT NA SUSP | -PR OV EST PT LEV 3<br>-IMMUN ADMIN IM/SQ/ID/PERC 1ST VAC ONLY<br>-IMMUNOTHERAPY MULTIPLE<br>-MAMMOGRAM SCREENING<br>-INFLUENZA VAC (FLU CLINIC ONLY) 3+YO PA |
| Abdominal pain, unspecified site (789.00) | -Abdominal pain, epigastric (789.06)<br>-Abdominal pain, other specified site (789.09)<br>-Esophageal reflux (530.81)<br>-Constipation, unspecified (564.00) | -CIPRO 500 MG PO TABS<br>-METRONIDAZOLE 500 MG PO TABS<br>-OMEPRAZOLE 20 MG PO CPDR<br>-HYDROCODONE-ACETAMINOPHEN 5-500 MG PO TABS<br>-ACIPHEX 20 MG PO TBEC | -US ABDOMEN COMPLETE<br>-GI, INT<br>-ABDOMEN/PELVIS CT PA<br>-PR ECG AND INTERPRETATION<br>-SERV PROV DURING REG SCHED EVE/WKEND/HOL HRS |
| (V72.3) | -Other general medical examination for administrative purposes (V70.3)<br>-Need for prophylactic vaccination and inoculation against tetanus-diphtheria [Td] (DT) (V06.5)<br>-Screening for unspecified condition (V82.9)<br>-Symptomatic menopausal or female climacteric states (627.2) | -ALBUTEROL 90 MCG/ACT IN AERS | -GYN CYTOLOGY (PAP) PA<br>-MAMMOGRAM SCREENING<br>-CHART ABSTRACTION SIMP (V70.3)<br>-PR OV EST PT LEV 2<br>-TD VAC 7+ YO (V06.5) IM |
| Other specified vaccinations against streptococcus pneumoniae [pneumococcus] (V03.82) | -Screening for malignant neoplasms of prostate (V76.44)<br>-Routine general medical examination at a health care | -ALBUTEROL SULFATE HFA 108 (90 BASE) MCG/ACT IN AERS | -PNEUMOCOCCAL VAC (V03.82) IM/SQ<br>-PNEUMOCOCCAL POLYSACC VAC 2+ YO (V03.82) IM/SQ |
| | facility (V70.0)<br>-Need for prophylactic vaccination and inoculation against tetanus-diphtheria [Td] (DT) (V06.5)<br>-Need for prophylactic vaccination and inoculation against other specified disease (V05.8) | | -IMMUN ADMIN IM/SQ/ID/PERC 1ST VAC ONLY<br>-INFLUENZA VAC 3+YR (V04.81) IM<br>-TDAP VAC 11-64 YO *ADACEL (V06.8) IM |
| Other psoriasis (696.1) | -Other dyschromia (709.09)<br>-Other atopic dermatitis and related conditions (691.8)<br>-Seborrheic dermatitis, unspecified (690.10)<br>-Unspecified pruritic disorder (698.9) | -CLOBETASOL PROPIONATE 0.05 % EX OINT<br>-CLOBETASOL PROPIONATE 0.05 % EX CREA<br>-FLUOCINONIDE 0.05 % EX CREA<br>-DESONIDE 0.05 % EX CREA<br>-TRIAMCINOLONE ACETONIDE 0.1 % EX CREA | -PR ULTRAVIOLET THERAPY<br>-DERMATOLOGY, INT<br>-PR BIOPSY OF SKIN/SQ/MUC MEMBR LESN SINGLE<br>-PR DESTR PRE-MALIG LESN 1ST<br>-DERMATOLOGY, INT |
| Other general medical examination for administrative purposes (V70.3) | -Routine general medical examination at a health care facility (V70.0)<br>-Need for prophylactic vaccination and inoculation against tetanus-diphtheria [Td] (DT) (V06.5)<br>-Screening for lipoid disorders (V77.91) | -ALBUTEROL 90 MCG/ACT IN AERS<br>-FLONASE 50 MCG/ACT NA SUSP | -CHART ABSTRACTION SIMP (V70.3)<br>-CHART ABSTRACTION INT (V70.3)<br>-PAMFONLINE ENROLLMENT<br>-MAMMOGRAM SCREENING<br>-PR CHART ABSTRACTION COMP (V70.3) |
| Migraine, unspecified, without mention of intractable migraine without mention of status migrainosus (346.90) | -Anxiety state, unspecified (300.00)<br>-Cervicalgia (723.1)<br>-Other screening mammogram (V76.12)<br>-Myalgia and myositis, unspecified (729.1) | -SUMATRIPTAN SUCCINATE 50 MG PO TABS<br>-HYDROCODONE-ACETAMINOPHEN 5-500 MG PO TABS<br>-PROMETHAZINE HCL 25 MG PO TABS<br>-IMITREX 100 MG PO TABS<br>-ZOLPIDEM TARTRATE 10 MG PO TABS | -MAMMO SCREENING BILATERAL DIGITAL<br>-GYN CYTOLOGY (PAP) PA<br>-PR OV EST PT LEV 3<br>-MAMMOGRAM SCREENING<br>-NEUROLOGY, INT |
| Routine general medical examination at a health care facility (V70.0) | -Other screening mammogram (V76.12)<br>-Special screening for malignant neoplasms of colon (V76.51)<br>-Screening for malignant neoplasms of prostate (V76.44)<br>-Need for prophylactic vaccination and inoculation against diphtheria-tetanus-pertussis, combined [DTP] [DTaP] (V06.1) | -PR OV EST PT LEV 3<br>-PR OV EST PT LEV 2<br>-IMMUN ADMIN IM/SQ/ID/PERC 1ST VAC ONLY<br>-TDAP VAC 11-64 YO *ADACEL (V06.8) IM<br>-GYN CYTOLOGY (PAP) PA | |
| Special screening for malignant neoplasms of colon (V76.51) | -Routine general medical examination at a health care facility (V70.0)<br>-Benign neoplasm of colon (211.3)<br>-Screening for unspecified condition (V82.9)<br>-Other screening mammogram (V76.12) | -PEG-KCL-NACL-NASULF-NA ASC-C 100 G PO SOLR | -GI, INT<br>-REFERRAL TO GI, INT<br>-PR OV EST PT LEV 3<br>-PR OV EST PT LEV 2<br>-IMMUN ADMIN IM/SQ/ID/PERC 1ST VAC ONLY |
| Care involving other physical therapy (V57.1) | -Pain in joint, shoulder region (719.41)<br>-Pain in joint, lower leg (719.46)<br>-Lumbago (724.2)<br>-Pain in joint, pelvic region and thigh (719.45) | -HYDROCODONE-ACETAMINOPHEN 5-500 MG PO TABS<br>-HYDROCODONE-ACETAMINOPHEN 10-325 MG PO TABS<br>-CYCLOBENZAPRINE HCL 10 MG PO TABS<br>-HYDROCODONE- | -PR PHYS THERAPY EVALUATION<br>-PR THERAPEUTIC EXERCISES<br>-PR MANUAL THER TECH 1+REGIONS EA 15 MIN<br>-PR ELECTRIC STIMULATION THERAPY<br>-REFERRAL TO PHYSICAL THERAPY, INT |
| | | ACETAMINOPHEN 5-325 MG PO TABS | |
| Rheumatoid arthritis (714.0) | -Unspecified inflammatory polyarthropathy (714.9)<br>-Arthropathy, unspecified, site unspecified (716.90)<br>-Sicca syndrome (710.2) | -FOLIC ACID 1 MG PO TABS<br>-METHOTREXATE 2.5 MG PO TABS<br>-PREDNISONE 5 MG PO TABS<br>-PREDNISONE 1 MG PO TABS<br>-HYDROXYCHLOROQUINE SULFATE 200 MG PO TABS | -RHEUMATOLOGY, INT<br>-METHYLPRED ACET 80MG *DEPO-MEDROL INJ<br>-TRIAMCINOLONE ACETONIDE *KENALOG INJ<br>-ASP/INJ INT JOINT/BURS/CYST<br>-PR METHLYPRED ACET (BU 21 - 40MG) *DEPO-MEDROL INJ |
| Osteoporosis, unspecified (733.00) | -Unspecified acquired hypothyroidism (244.9)<br>-Osteoarthrosis, unspecified whether generalized or localized, site unspecified (715.90)<br>-Benign essential hypertension (401.1)<br>-Unspecified essential hypertension (401.9) | -FOSAMAX 70 MG PO TABS<br>-ALENDRONATE SODIUM 70 MG PO TABS<br>-HYDROCHLOROTHIAZIDE 25 MG PO TABS<br>-ACTONEL 35 MG PO TABS | -DEXA BONE DENSITY AXIAL SKELETON HIPS/PELV/SPINE<br>-BONE DENSITY (DEXA)<br>-DEXA BONE DENSITY STUDY 1+ SITES AXIAL SKEL (HIP/PELVIS/SPINE)<br>-INFLUENZA VAC 3+YR (V04.81) IM<br>-MAMMOGRAM SCREENING |
| Other malaise and fatigue (780.79) | -Routine general medical examination at a health care facility (V70.0)<br>-Dizziness and giddiness (780.4)<br>-Other respiratory abnormalities (786.09)<br>-Screening for unspecified condition (V82.9) | -AZITHROMYCIN 250 MG PO TABS<br>-FLUTICASONE PROPIONATE 50 MCG/ACT NA SUSP | -PR ECG AND INTERPRETATION<br>-CHEST PA & LATERAL<br>-PR OV EST PT LEV 3<br>-EKG<br>-TDAP VAC 11-64 YO *ADACEL (V06.8) IM |
| Long-term (current) use of anticoagulants (V58.61) | -Atrial fibrillation (427.31)<br>-Congestive heart failure, unspecified (428.0)<br>-Coronary atherosclerosis of unspecified type of vessel, native or graft (414.00) | -WARFARIN SODIUM 5 MG PO TABS<br>-COUMADIN 5 MG PO TABS<br>-WARFARIN SODIUM 2.5 MG PO TABS<br>-ATENOLOL 50 MG PO TABS<br>-WARFARIN SODIUM 2 MG PO TABS | -PR OV EST PT MIN SERV<br>-PR PROTHROMBIN TIME<br>-DEBRIDEMENT OF NAILS, 6 OR MORE<br>-PR ECG AND INTERPRETATION<br>-EKG |
| Spinal stenosis, lumbar region, without neurogenic claudication (724.02) | -Degeneration of lumbar or lumbosacral intervertebral disc (722.52)<br>-Lumbosacral spondylosis without myelopathy (721.3)<br>-Displacement of lumbar intervertebral disc without myelopathy (722.10)<br>-Lumbago (724.2) | -HYDROCODONE-ACETAMINOPHEN 10-325 MG PO TABS<br>-GABAPENTIN 300 MG PO CAPS<br>-HYDROCODONE-ACETAMINOPHEN 5-500 MG PO TABS<br>-TRAMADOL HCL 50 MG PO TABS<br>-PREDNISONE 20 MG PO TABS | -LUMBAR SPINE MRI W/O CONTR<br>-LS SPINE 2 VIEWS<br>-PHYSICAL MEDICINE, INT<br>-MRI LUMBAR SPINE WO CONTRAST<br>-PHYSICAL MEDICINE, INT |
| Unspecified hemorrhoids without mention of complication (455.6) | -Internal hemorrhoids without mention of complication (455.0)<br>-Benign neoplasm of colon (211.3)<br>-Anal or rectal pain (569.42)<br>-Personal history of colonic polyps (V12.72) | -HYDROCORTISONE ACETATE 25 MG PR SUPP<br>-ANUSOL-HC 25 MG PR SUPP<br>-HYDROCORTISONE 2.5 % PR CREA<br>-ANUSOL-HC 2.5 % PR CREA<br>-PEG-KCL-NACL-NASULF-NA ASC-C 100 G PO SOLR | -ANOSCOPY<br>-SURGERY, INT<br>-PR COLONOSCOPY W/BIOPSY(S)<br>-LIGATION OF HEMORRHOID(S)<br>-GI, INT |
| Dermatophytosis of nail (110.1) | -Dermatophytosis of foot (110.4)<br>-Ingrowing nail (703.0)<br>-Keratoderma, acquired (701.1)<br>-Peripheral vascular disease, unspecified (443.9) | -ECONAZOLE NITRATE 1 % EX CREA<br>-TERBINAFINE HCL 250 MG PO TABS<br>-TRIAMCINOLONE ACETONIDE 0.1 % EX CREA | -DEBRIDEMENT OF NAILS, 6 OR MORE<br>-TRIM SKIN LESIONS, 2 TO 4<br>-PARE CORN/CALLUS, SINGLE LESION<br>-PODIATRY, INT |
| | | -CEPHALEXIN 500 MG PO CAPS | -PR DESTR PRE-MALIG LESN 1ST |
| Pain in joint, lower leg (719.46) | -Chondromalacia of patella (717.7)<br>-Tear of medial cartilage or meniscus of knee, current (836.0)<br>-Pain in limb (729.5)<br>-Pain in joint, pelvic region and thigh (719.45) | -HYDROCODONE-ACETAMINOPHEN 5-500 MG PO TABS<br>-VICODIN 5-500 MG PO TABS<br>-IBUPROFEN 600 MG PO TABS | -ASP/INJ MAJOR JOINT/BURS/CYST<br>-REFERRAL TO ORTHOPEDICS, INT<br>-TRIAMCINOLONE ACETONIDE *KENALOG INJ<br>-LOWER EXTREMITY JOINT MRI W/O CONTR<br>-ORTHOPEDICS, INT |
| Pain in joint, pelvic region and thigh (719.45) | -Lumbago (724.2)<br>-Sciatica (724.3)<br>-Backache, unspecified (724.5)<br>-Pain in joint, lower leg (719.46) | -HYDROCODONE-ACETAMINOPHEN 5-500 MG PO TABS<br>-VICODIN 5-500 MG PO TABS | -PELVIS LIMITED<br>-LS SPINE 2 VIEWS<br>-XR PELVIS 1 OR 2 VIEWS<br>-REFERRAL TO ORTHOPEDICS, INT<br>-ASP/INJ MAJOR JOINT/BURS/CYST |
| Anxiety state, unspecified (300.00) | -Insomnia, unspecified (780.52)<br>-Generalized anxiety disorder (300.02)<br>-Other malaise and fatigue (780.79)<br>-Headache (784.0) | -LORAZEPAM 0.5 MG PO TABS<br>-ZOLPIDEM TARTRATE 10 MG PO TABS<br>-LORAZEPAM 1 MG PO TABS<br>-ALPRAZOLAM 0.25 MG PO TABS<br>-CLONAZEPAM 0.5 MG PO TABS | -MAMMO SCREENING BILATERAL DIGITAL<br>-EKG<br>-PR ECG AND INTERPRETATION<br>-PR INFLUENZA VACC PRES FREE 3+ YO 0.5ML IM<br>-TDAP VAC 11-64 YO *ADACEL (V06.8) IM |
| Generalized anxiety disorder (300.02) | -Depressive disorder, not elsewhere classified (311)<br>-Insomnia, unspecified (780.52)<br>-Irritable bowel syndrome (564.1)<br>-Sleep disturbance, unspecified (780.50) | -LORAZEPAM 0.5 MG PO TABS<br>-CLONAZEPAM 0.5 MG PO TABS<br>-TRAZODONE HCL 50 MG PO TABS<br>-LORAZEPAM 1 MG PO TABS<br>-AMBIEN 10 MG PO TABS | -MAMMOGRAM SCREENING<br>-GI, INT<br>-CHART ABSTRACTION INT (V70.3) |
| Cervicalgia (723.1) | -Lumbago (724.2)<br>-Backache, unspecified (724.5)<br>-Headache (784.0)<br>-Brachial neuritis or radiculitis NOS (723.4) | -CYCLOBENZAPRINE HCL 10 MG PO TABS<br>-HYDROCODONE-ACETAMINOPHEN 5-500 MG PO TABS<br>-FLEXERIL 10 MG PO TABS<br>-NAPROXEN 500 MG PO TABS<br>-CARISOPRODOL 350 MG PO TABS | -PHYSICAL MEDICINE, INT<br>-SPINE CERVICAL COMPLETE<br>-PHYSICAL THERAPY, INT<br>-PHYSICAL THERAPY, EXT<br>-PR THERAPEUTIC EXERCISES |
| Impaired fasting glucose (790.21) | -Routine general medical examination at a health care facility (V70.0)<br>-Screening for malignant neoplasms of prostate (V76.44)<br>-Other and unspecified hyperlipidemia (272.4)<br>-Overweight (278.02) | -SIMVASTATIN 20 MG PO TABS<br>-HYDROCHLOROTHIAZIDE 25 MG PO TABS<br>-SIMVASTATIN 40 MG PO TABS | -PR OV EST PT LEV 3<br>-TDAP VAC 11-64 YO *ADACEL (V06.8) IM<br>-PR OV EST PT LEV 2<br>-NUTRITION, INT<br>-PR INFLUENZA VACC PRES FREE 3+ YO 0.5ML IM |
| Actinic keratosis (702.0) | -Other chronic dermatitis due to solar radiation (692.74)<br>-Other seborrheic keratosis (702.19)<br>-Benign neoplasm of other specified sites of skin (216.8)<br>-Neoplasm of uncertain behavior of skin (238.2) | -TRIAMCINOLONE ACETONIDE 0.1 % EX CREA<br>-FLUOCINONIDE 0.05 % EX CREA<br>-GLASSES | -PR DESTR PRE-MALIG LESN 1ST<br>-PR DESTR PRE-MALIG LESN 2-14<br>-PR BIOPSY OF SKIN/SQ/MUC MEMBR LESN SINGLE<br>-SURGICAL PATHOLOGY PA<br>-PR OV EST PT LEV 3 |